\documentclass[11pt]{article}

\usepackage[preprint]{acl}

\usepackage{times}
\usepackage{latexsym}
\usepackage[T1]{fontenc}

\usepackage[utf8]{inputenc}

\usepackage{microtype}

\usepackage{inconsolata}

\usepackage{graphicx}
\usepackage{booktabs}
\usepackage{multirow}
\usepackage{amsmath}
\usepackage{amssymb}
\usepackage{xcolor}
\usepackage{colortbl}
\definecolor{tbblue}{RGB}{0,120,215}
\definecolor{tbred}{RGB}{200,50,50}
\definecolor{figarrowblue}{RGB}{0,104,183}
\definecolor{figarroworange}{RGB}{230,126,34}
\definecolor{figarrowred}{RGB}{225,60,55}
\usepackage{makecell}
\usepackage{subcaption}
\usepackage{placeins}
\usepackage[most]{tcolorbox}
\usepackage{enumitem}
\definecolor{promptbg}{RGB}{245,248,255}
\definecolor{promptframe}{RGB}{70,130,200}
\definecolor{statebg}{RGB}{245,255,248}
\definecolor{stateframe}{RGB}{60,160,100}
\definecolor{factbg}{RGB}{255,248,240}
\definecolor{factframe}{RGB}{200,130,50}

\title{Intent-Driven Situation Tracking for User-Centric Multi-Turn Agents}

\author{
  Meiling Tao\textsuperscript{1}, \
  Yiling Tao\textsuperscript{2}, \
  Peng Wang\textsuperscript{1}\thanks{\ Corresponding author.} \\[2pt]
  \textsuperscript{1}University of Electronic Science and Technology of China \\
  \textsuperscript{2}Shenzhen International Graduate School, Tsinghua University \\[2pt]
  \texttt{meilingtao.cs@gmail.com}, \texttt{p.wang6@hotmail.com} \\
}

\begin{document}
\maketitle
\begin{abstract}
User-centric multi-turn agents must act on an evolving task situation shaped by changing user intents, accumulated tool-grounded facts, missing information, and execution constraints. Existing context-management methods improve the use of past interaction history, but rarely maintain an explicit situation state that separates grounded facts from task-state judgments. As a result, agents often need to infer fine-grained attributes, task dependencies, and constraint satisfaction implicitly from dialogue traces. We propose Intent-Driven Situation States (IDSS), a training-free framework that maintains an explicit situation state alongside the dialogue. IDSS parses tool returns into provenance-aware entities and attributes, tracks user intents, required variables, constraints, and execution status, and propagates new facts to task constraints to update action executability. This allows agents to avoid infeasible actions, advance dependent goals, and reuse relevant information without repeatedly searching raw history. Experiments on three interactive benchmarks across eight LLMs show that IDSS improves task completion, preference elicitation, and interaction efficiency, with clear gains on tasks involving multi-entity coordination, evolving user constraints, and constraint-aware replanning. Ablations and error analyses show that these improvements come from the interaction between fact persistence, intent-centered state tracking, and constraint modeling. These results suggest that explicit situation tracking offers an effective alternative to history-centric context management for reliable user-centric multi-turn agents.

\end{abstract}

\section{Introduction}
\label{sec:introduction}

\begin{figure*}[!t]
    \centering
    \includegraphics[width=\textwidth]{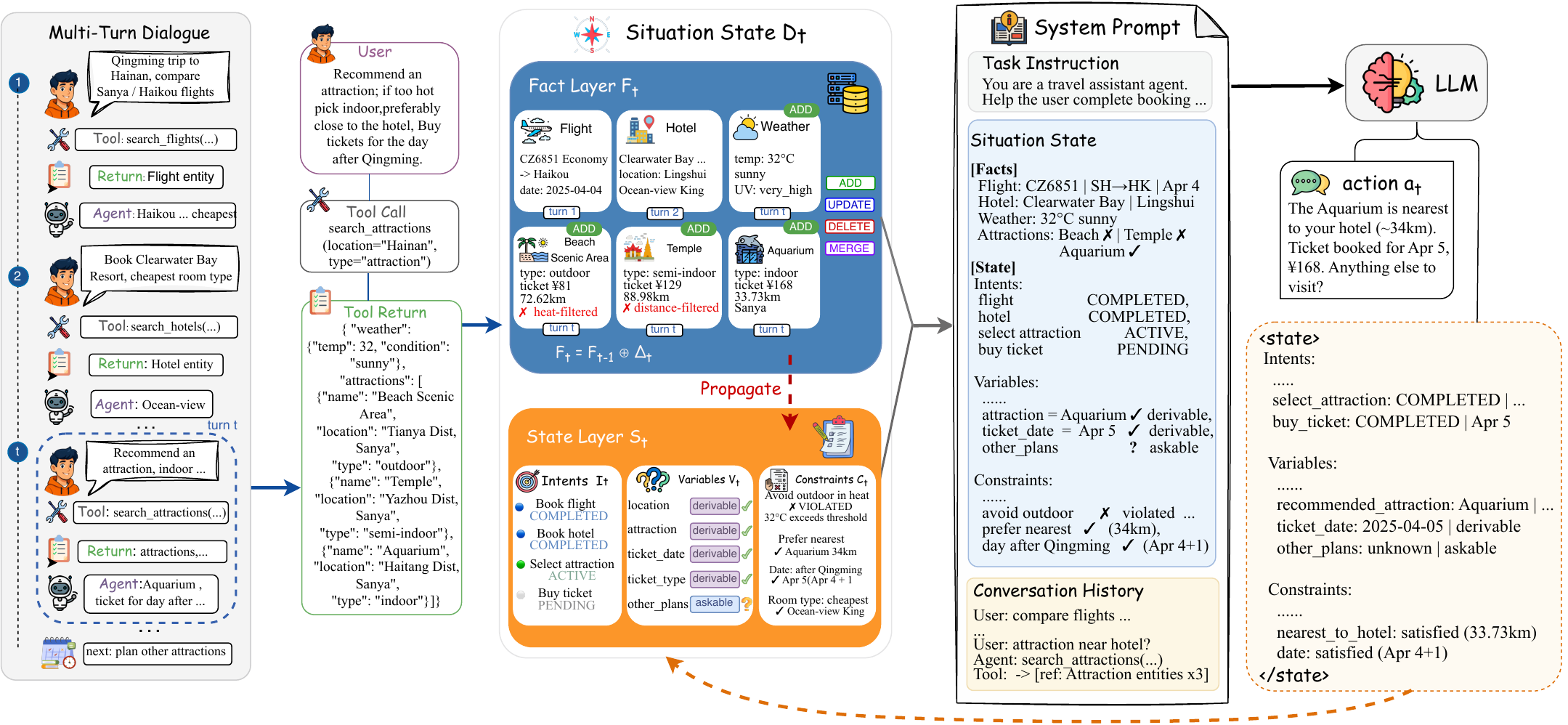}
    \caption{Overview of the IDSS framework. At each turn, the agent receives a user message or tool return: the fact layer converts tool returns into structured entities via deterministic parsing, while the state layer is updated by the agent during reasoning to reflect intents, variables, and constraints. New facts trigger cross-layer constraint propagation (\textcolor{figarrowred}{red arrows}), which may block infeasible intents and activate subsequent ones. Left: multi-turn dialogue timeline. Right: prompt assembly and LLM output. \textcolor{figarrowblue}{Blue arrows} indicate tool returns parsed into the fact layer. \textcolor{figarroworange}{Orange arrows} indicate the updated state block being fed back into the next prompt.}
    \label{fig:framework}
\end{figure*}

LLM-based agents are increasingly used in user-facing applications such as customer service, travel planning, and life services~\citep{schick2023toolformer,qin2024toolllm,patil2024gorilla}.
In these scenarios, agents no longer execute isolated instructions~\citep{wei2022chain,yao2023tree}, but instead engage in sustained interactions where they must understand evolving user goals, call external tools, follow rules, and decide when to ask, retrieve, execute, or finish.
User-centric multi-turn tasks therefore place stronger demands on state maintenance than single-shot tool use or single-goal long-horizon planning.

The difficulty comes from the evolving nature of user interactions.
Users may introduce multiple goals in one conversation, and these goals can depend on or conflict with one another.
Preferences are often released gradually rather than stated upfront.
Tool returns may revise previously assumed facts, and newly observed facts may make an earlier action path infeasible.
Completing such tasks requires agents to track the current task situation: which facts are grounded, which intents are active or completed, which variables are still missing, and which constraints are satisfied or violated.
Without such tracking, an agent must repeatedly reconstruct the current situation from a growing dialogue history, which can lead to fact omission, intent drift, premature execution, and constraint violations.
Recent user-interaction benchmarks~\citep{yao2024tau,he2025vitabench,qian2025userbench} show that even strong models struggle with these issues in realistic multi-turn settings.

Prior work addresses long interactions mainly through external memory or context compression~\citep{packer2023memgpt,chhikara2025mem0,rasmussen2025zep,wu2025resum,ye2025agentfold,su2026u}.
External memory systems make interaction information retrievable beyond the immediate context window, while compression, summarization, and folding methods shorten or restructure previous turns to fit the prompt.
These approaches reduce information access and context-length pressure, but the agent must still infer the current task situation from retrieved memories, summaries, or compressed traces.
Grounded facts are rarely organized separately from task-state judgments such as active intents, missing variables, and constraints, leaving fine-grained attributes, task dependencies, and constraint satisfaction to implicit model inference.

To address this gap, we propose \textbf{Intent-Driven Situation States} (IDSS), a training-free framework for situation tracking in user-centric multi-turn agents.
IDSS maintains an explicit \textit{situation state} alongside the dialogue history.
The fact layer parses tool returns into provenance-aware entities and attributes, while the state layer tracks user intents, required variables, constraints, and execution status.
Cross-layer constraint propagation links newly observed facts to task constraints, allowing the agent to block infeasible intents, activate subsequent goals, and reuse relevant information without repeatedly searching raw history.
At each turn, the situation state is rendered into the prompt together with a compact residual dialogue history, providing the agent with a decision-oriented view of the current task situation, as shown in Figure~\ref{fig:framework}.

We evaluate IDSS on $\tau$-bench, VitaBench, and UserBench across eight LLMs. Our contributions are as follows:

\begin{itemize}
    \item We formulate user-centric multi-turn interaction as intent-driven situation tracking and propose IDSS, which maintains explicit situation states separating grounded facts from task-state judgments.
    \item We design a dual-layer update mechanism requiring no additional LLM calls: the fact layer updates incrementally through deterministic parsing, and the state layer is maintained during action generation to track intents, variables, constraints, and execution status.
    \item Experiments on three interactive benchmarks across eight LLMs show that IDSS improves task completion, preference elicitation, and interaction efficiency, with the clearest gains on tasks requiring multi-entity coordination and evolving constraints.
\end{itemize}

\section{Related Work}
\label{sec:related}

\subsection{User-Centric Multi-Turn Agents}

Recent benchmarks show that single-turn performance does not reliably predict multi-turn capabilities~\citep{xi2025rise,wang2024survey}. $\tau$-bench and $\tau^2$-bench~\citep{yao2024tau,barres2025tau} test policy-following and tool-use consistency in customer-service settings, VitaBench~\citep{he2025vitabench} stresses temporal and spatial reasoning and shifting intents across life-service domains, and UserBench~\citep{qian2025userbench} focuses on incremental preference elicitation. Even strong models still struggle on these benchmarks, revealing systematic challenges in information persistence, intent tracking, and holistic task planning as conversations grow longer. On the method side, \citet{sun2025training} train proactive interaction agents via instruction tuning, UserRL~\citep{qian2025userrl} applies reinforcement learning with multi-turn credit assignment, and \citet{suri2025structured} model interactions as a POMDP for structured clarification. While effective, these approaches often rely on dedicated supervision, training procedures, or task-specific assumptions, which may limit cross-domain generalization. IDSS requires no additional training and instead improves multi-turn interaction by restructuring context into an explicit situation state.

\subsection{Agent Context Management}

Existing agent context-management methods often address long-interaction challenges through external memory or in-context compression~\citep{zhang2025survey}. External-memory systems, such as MemGPT, Mem0, Graphiti, and Generative Agents~\citep{packer2023memgpt,chhikara2025mem0,rasmussen2025zep,park2023generative}, improve retention through working-memory management, long-term fact consolidation, temporal knowledge graphs, or reflection-based retrieval, but do not necessarily maintain an explicit current task situation. In-context compression methods such as ReSum~\citep{wu2025resum} and IterResearch~\citep{chen2025iterresearch} condense interaction histories into compact reasoning states, while LLMLingua~\citep{jiang2023llmlingua,jiang2024longllmlingua} and AutoCompressors~\citep{chevalier2023adapting} prune token-level redundancy. AgentFold~\citep{ye2025agentfold} and U-Fold~\citep{su2026u} further perform more semantic compression through sub-task folding or intent-aware extraction.
However, they rarely separate tool-grounded facts from task-state judgments such as intents, missing variables, and constraints, which IDSS maintains in an explicit situation state for decision-making.

\subsection{State Tracking}

Dialogue state tracking (DST) maintains slot-value pairs within predefined ontologies~\citep{budzianowski2018multiwoz,heck2020trippy}, with recent extensions using LLM function calling~\citep{li2024large} or knowledge-graph reasoning~\citep{pan2024unifying,jiang2023structgpt}. However, DST is not designed for the open-ended entities, dynamic constraints, and multi-intent dependencies in tool-augmented agent settings. In the agent domain, StateAct~\citep{rozanov2025stateact} proposes chain-of-states to track environment state at every step, but does not explicitly separate tool-grounded facts from user task goals, nor does it model constraint dependencies. PABU~\citep{jiang2026pabu} maintains a belief state via progress prediction and selective history retention, yet its linear progress assumption may be insufficient for non-linear multi-intent dependencies, and it requires fine-tuning on curated trajectories. IDSS instead separates tool-grounded facts from task-state judgments and links them through constraint propagation, allowing non-linear intent dependencies and blocking conditions to be updated across turns.

\section{Method}
\label{sec:method}

\subsection{Problem Formulation}
\label{sec:formulation}

Consider an LLM agent that assists a user through multi-turn interactions. At each turn $t$, the agent receives an input $x_t$, which can be either a user message or a tool return, and generates an output $a_t$, such as a tool call or a user-facing response. Let $H_t$ denote the interaction context available when generating $a_t$:
\begin{equation}
H_t = (x_1, a_1, x_2, a_2, \ldots, a_{t-1}, x_t).
\end{equation}
Given $H_t$ and the available tool set $\mathcal{A}$, the agent follows a policy $\pi$:
\begin{equation}
a_t = \pi(H_t, \mathcal{A}).
\end{equation}


Under the standard ReAct paradigm, $H_t$ is maintained as a linear sequence of user messages, tool returns, and agent outputs. As interactions grow, this linear history becomes inadequate for user-centric multi-turn tasks: relevant information may be truncated or compressed, tool-grounded attributes may be buried in long traces, and preferences, tool preconditions, and rules may be scattered across turns. The agent must repeatedly reconstruct the current task situation from raw history, which can lead to fact omission, intent drift, premature execution, and constraint violations.

\subsection{Intent-Driven Situation Tracking}
\label{sec:overview}

To address these limitations, we propose \textbf{Intent-Driven Situation States (IDSS)}, a training-free framework that maintains an explicit situation state alongside the dialogue history. Rather than relying on the agent to infer the current situation from a long, mixed sequence of user messages, tool calls, and tool returns, IDSS organizes multi-turn interaction around evolving user intents, while grounding decisions in structured facts, required variables, and execution constraints.

Formally, IDSS replaces the history-only policy with a situation-augmented policy:
\begin{equation}
a_t = \pi(D_t, H_t^{\mathrm{comp}}, \mathcal{A}),
\end{equation}
where $D_t$ denotes the structured situation state maintained by IDSS, and $H_t^{\mathrm{comp}}$ is the compact residual history obtained from $H_t$ after state-covered information is removed or replaced with references to $D_t$. In practice, $H_t^{\mathrm{comp}}$ retains recent user-agent turns while replacing parsed tool outputs with references to entries in the fact layer. Through this separation, the agent acts on an explicit representation of the current task situation together with a lightweight dialogue trace, rather than searching through unorganized raw history.

As shown in Figure~\ref{fig:framework}, IDSS consists of three components: a fact layer for tool-grounded information, a state layer for user-task progress, and a cross-layer propagation step that aligns task progress with newly observed facts. Formally, the situation state is defined as:
\begin{equation}
D_t = (F_t, S_t),
\end{equation}
where $F_t$ stores structured entities and attributes extracted from tool returns, and $S_t$ tracks user intents, required variables, constraints, and execution status.

\subsubsection{Fact Layer}
\label{sec:fact_layer}

Tool returns across turns often describe interrelated entities, such as flights, hotels, and orders, together with attributes such as price, location, date, and status. The fact layer organizes these observations into a structured entity store, enabling cross-entity and cross-turn querying without tracing back through raw conversation history.

\paragraph{Extraction and Update.}
The fact layer is defined as:
\begin{equation}
F_t = \{e_i = (\text{type}_i, \text{id}_i, \text{attrs}_i, \text{src}_i)\},
\end{equation}
where $\text{type}_i$ denotes the entity type, $\text{id}_i$ the entity identifier, $\text{attrs}_i$ the attribute set, and $\text{src}_i$ the provenance source.

Let $o_t$ denote a tool-return observation when $x_t$ is produced by a tool. A deterministic parser converts $o_t$ into an operation sequence:
\begin{equation}
\Delta_t = \mathrm{Parse}(o_t), \quad \Delta_t = [\delta_1, \delta_2, \ldots],
\end{equation}
where each $\delta_i$ is an atomic operation. The fact layer is then updated incrementally:
\begin{equation}
F_t = F_{t-1} \oplus \Delta_t.
\end{equation}
Operations include \texttt{ADD}, \texttt{UPDATE} for attribute-level merging, \texttt{DELETE} for marking invalid entities, and \texttt{MERGE} for combining duplicates.

\paragraph{Conflict Resolution and Compression.}
When the same attribute receives conflicting values, the fact layer resolves them by source credibility. Tool-observed attributes are prioritized over agent-inferred ones, while user preferences and constraints are maintained in the state layer. Within the same source type, newer observations update older values. This prevents hallucinated or stale information from polluting the fact store.

After entity extraction, corresponding tool returns in the conversation history are replaced with brief fact-layer references. This keeps the prompt compact while preserving factual information in an explicit and accessible form.

\subsubsection{State Layer}
\label{sec:state_layer}

While the fact layer stores tool-grounded observations, the state layer tracks task-state judgments that evolve across turns. It is organized around user intents and records intent progress, dependencies, required variables, and constraints. The agent is instructed to prepend a \texttt{<state>} block to each response, listing intents, confirmed conclusions, and missing information. This block is parsed into $S_t$, removed before user delivery, and re-injected into the next prompt:
\begin{equation}
S_t = (I_t, V_t, C_t),
\end{equation}
where $I_t$ is the intent set, $V_t$ the variable set, and $C_t$ the constraint set.

The state layer also guides normal completion decisions: unless the user explicitly stops or an external turn limit is reached, the agent treats the current task as complete when all user-introduced intents are \texttt{COMPLETED} or \texttt{BLOCKED} and no required variables remain unknown for executable intents. Otherwise, remaining \texttt{PENDING} intents or unresolved variables guide the next question, retrieval, or tool action.

\paragraph{Intent Tracking.}
The agent identifies explicit or implicit task goals from user messages and records them as intents. New user requests are appended to the intent set. Each intent is assigned one of four statuses: \texttt{active}, \texttt{pending}, \texttt{completed}, or \texttt{blocked}.

Intent dependencies are inferred from task semantics and execution preconditions. If an intent requires another intent, user confirmation, tool-returned information, or satisfied constraints before execution, it is linked to the corresponding prerequisite. Once dependencies are completed, required variables are known, and relevant constraints are not violated, the intent becomes \texttt{active}; after execution, it becomes \texttt{completed}. Intent status therefore guides the next action by identifying executable goals, information gaps, and infeasible paths.

\paragraph{Variable Tracking.}
The variable set describes information slots required by current or pending intents. Each variable is annotated as \texttt{askable} if it should be obtained from the user, \texttt{retrievable} if it requires tool invocation, or \texttt{derivable} if it can be inferred from existing facts. Variables are also marked as known or unknown, helping the agent decide whether to ask, retrieve, or derive information.

\paragraph{Constraint Modeling.}
The constraint set records conditions that actions must satisfy, including user preferences, tool preconditions, and rules. Each constraint is represented as a short rule description with grounded arguments when available, such as an entity attribute, operator, and required value, and its status is marked as \texttt{satisfied}, \texttt{unsatisfied}, or \texttt{violated}. For example, in the travel-planning case in Figure~\ref{fig:framework}, an attraction-selection intent may include constraints such as \texttt{nearest\_to\_hotel} or \texttt{avoid\_outdoor}, whose status is updated when hotel-location or weather facts become available. When a constraint becomes \texttt{violated}, the associated intent is marked as \texttt{blocked}, preventing repeated attempts at infeasible actions.

\subsection{Cross-Layer Constraint Propagation}
\label{sec:propagation}

The fact and state layers are connected through cross-layer constraint propagation. Whenever the fact layer updates entity attributes, the system re-evaluates grounded constraints in $C_t$ through deterministic attribute comparison. Constraints without sufficient grounding remain \texttt{unsatisfied} rather than being incorrectly marked as \texttt{violated}. When a constraint is violated, the system blocks the associated intent and records the factual basis and constraint chain behind the decision.

This mechanism allows the agent to understand why a path is infeasible and derive alternatives instead of retrying failed operations. For example, a newly observed fare-class attribute may violate a flight-modification constraint, block the modification intent, and activate a cancellation-and-rebooking intent while preserving the user's original requirements. Figure~\ref{fig:case} illustrates this mechanism on a concrete $\tau$-bench Airline task.

\section{Experiments}
\label{sec:experiments}

\begin{table*}[!t]
    \centering
    \small
    \caption{Main results across three benchmarks with eight models. \textbf{Bold} indicates best, \underline{underline} indicates second best within each model group. Row background reflects improvement over the ReAct baseline: deeper \colorbox{tbblue!35}{\strut blue} indicates larger gains, deeper \colorbox{tbred!24}{\strut red} indicates larger degradation.}
    \label{tab:main_results}
    \resizebox{0.85\textwidth}{!}{%
    \begin{tabular}{cl|cccc|cccc|ccc}
    \toprule
    \multirow{3}{*}{\textbf{Model}} & \multirow{3}{*}{\textbf{Method}} 
    & \multicolumn{4}{c|}{\textbf{$\tau$-bench}} 
    & \multicolumn{4}{c|}{\textbf{VitaBench (Avg@4)}} 
    & \multicolumn{3}{c}{\textbf{UserBench}} \\
    \cmidrule(lr){3-6} \cmidrule(lr){7-10} \cmidrule(lr){11-13}
    & & \multicolumn{2}{c}{Retail} & \multicolumn{2}{c|}{Airline} 
    & \multirow{2}{*}{Delivery} & \multirow{2}{*}{In-store} & \multirow{2}{*}{OTA} & \multirow{2}{*}{Cross} 
    & \multirow{2}{*}{Score} & \multirow{2}{*}{CER} & \multirow{2}{*}{PE (\%)} \\
    & & Avg@4 & pass\textsuperscript{4} & Avg@4 & pass\textsuperscript{4} & & & & & & & \\
    \midrule
    \multirow{6}{*}{\textit{GPT-5.5}} & ReAct        & 77.5 & 46.0 & 63.5 & 28.5 & 52.0 & 61.5 & 34.0 & 22.5 & 0.368 & 0.405 & 28.5 \\
    & StateAct     & \cellcolor{tbblue!8}78.2 & \cellcolor{tbblue!8}47.0 & \cellcolor{tbblue!8}64.2 & \cellcolor{tbblue!8}29.5 & \cellcolor{tbblue!8}52.5 & \cellcolor{tbblue!8}62.0 & \cellcolor{tbblue!16}35.5 & \cellcolor{tbblue!16}23.5 & \cellcolor{tbblue!8}0.378 & \cellcolor{tbblue!8}0.414 & \cellcolor{tbblue!8}29.0 \\
    & ReSum        & \cellcolor{tbred!30}52.0 & \cellcolor{tbred!30}28.5 & \cellcolor{tbred!24}48.5 & \cellcolor{tbred!24}22.0 & \cellcolor{tbred!18}44.5 & \cellcolor{tbred!24}52.0 & \cellcolor{tbred!18}30.2 & \cellcolor{tbblue!36}25.5 & \cellcolor{tbblue!16}0.380 & \cellcolor{tbblue!16}0.418 & \cellcolor{tbblue!16}29.8 \\
    & IterResearch & \cellcolor{tbred!18}70.5 & \cellcolor{tbred!18}38.5 & \cellcolor{tbred!24}52.5 & \cellcolor{tbred!18}23.5 & \cellcolor{tbred!30}30.0 & \cellcolor{tbred!30}44.5 & \cellcolor{tbred!30}20.0 & \cellcolor{tbred!30}12.0 & \cellcolor{tbblue!16}0.385 & \cellcolor{tbblue!16}0.422 & \cellcolor{tbblue!16}30.2 \\
    & U-Fold       & \cellcolor{tbblue!8}\underline{79.0} & \cellcolor{tbblue!16}\underline{48.5} & \cellcolor{tbblue!8}\underline{65.0} & \cellcolor{tbblue!16}\underline{30.5} & \cellcolor{tbblue!26}\textbf{57.0} & \cellcolor{tbblue!16}\textbf{64.0} & \cellcolor{tbblue!26}\underline{38.0} & \cellcolor{tbblue!46}\underline{27.0} & \cellcolor{tbblue!26}\underline{0.395} & \cellcolor{tbblue!26}\underline{0.435} & \cellcolor{tbblue!26}\underline{31.5} \\
    & IDSS & \cellcolor{tbblue!16}\textbf{80.5} & \cellcolor{tbblue!26}\textbf{51.0} & \cellcolor{tbblue!16}\textbf{66.5} & \cellcolor{tbblue!36}\textbf{33.0} & \cellcolor{tbblue!26}\underline{56.5} & \cellcolor{tbblue!16}\underline{63.5} & \cellcolor{tbblue!36}\textbf{40.5} & \cellcolor{tbblue!46}\textbf{29.8} & \cellcolor{tbblue!36}\textbf{0.415} & \cellcolor{tbblue!36}\textbf{0.455} & \cellcolor{tbblue!36}\textbf{33.5} \\
    \midrule
    \multirow{6}{*}{\textit{GPT-4.1}} & ReAct        & 71.8 & 38.0 & 56.0 & 22.0 & 46.0 & 56.2 & 28.5 & 17.8 & 0.330 & 0.365 & 24.2 \\
    & StateAct     & \cellcolor{tbblue!8}72.6 & \cellcolor{tbblue!16}39.2 & \cellcolor{tbblue!8}56.8 & \cellcolor{tbblue!16}23.2 & \cellcolor{tbblue!8}46.3 & \cellcolor{tbblue!8}56.4 & \cellcolor{tbblue!16}30.0 & \cellcolor{tbblue!16}18.2 & \cellcolor{tbblue!16}0.340 & \cellcolor{tbblue!8}0.374 & \cellcolor{tbblue!8}24.8 \\
    & ReSum        & \cellcolor{tbred!30}46.5 & \cellcolor{tbred!30}24.0 & \cellcolor{tbred!24}42.5 & \cellcolor{tbred!24}18.0 & \cellcolor{tbred!18}39.5 & \cellcolor{tbred!24}47.0 & \cellcolor{tbred!18}26.2 & \cellcolor{tbblue!36}20.8 & \cellcolor{tbblue!16}0.342 & \cellcolor{tbblue!16}0.378 & \cellcolor{tbblue!16}25.5 \\
    & IterResearch & \cellcolor{tbred!18}64.8 & \cellcolor{tbred!18}32.5 & \cellcolor{tbred!24}46.5 & \cellcolor{tbred!18}19.5 & \cellcolor{tbred!30}25.5 & \cellcolor{tbred!30}39.0 & \cellcolor{tbred!30}16.5 & \cellcolor{tbred!30}9.2  & \cellcolor{tbblue!16}0.348 & \cellcolor{tbblue!16}0.384 & \cellcolor{tbblue!16}25.8 \\
    & U-Fold       & \cellcolor{tbblue!8}\underline{73.5} & \cellcolor{tbblue!16}\underline{40.5} & \cellcolor{tbblue!8}\underline{57.5} & \cellcolor{tbblue!26}\underline{24.5} & \cellcolor{tbblue!26}\textbf{51.5} & \cellcolor{tbblue!16}\textbf{58.5} & \cellcolor{tbblue!26}\underline{32.5} & \cellcolor{tbblue!46}\underline{22.0} & \cellcolor{tbblue!26}\underline{0.360} & \cellcolor{tbblue!26}\underline{0.398} & \cellcolor{tbblue!26}\underline{27.5} \\
    & IDSS & \cellcolor{tbblue!16}\textbf{75.0} & \cellcolor{tbblue!36}\textbf{43.0} & \cellcolor{tbblue!16}\textbf{59.0} & \cellcolor{tbblue!46}\textbf{26.5} & \cellcolor{tbblue!26}\underline{51.2} & \cellcolor{tbblue!16}\underline{58.0} & \cellcolor{tbblue!46}\textbf{34.8} & \cellcolor{tbblue!46}\textbf{24.8} & \cellcolor{tbblue!36}\textbf{0.382} & \cellcolor{tbblue!36}\textbf{0.418} & \cellcolor{tbblue!36}\textbf{29.5} \\
    \midrule
    \multirow{6}{*}{\textit{GPT-4o}} & ReAct        & 68.5 & 34.0 & 52.0 & 18.5 & 42.0 & 52.3 & 26.0 & 14.5 & 0.312 & 0.345 & 22.5 \\
    & StateAct     & \cellcolor{tbblue!8}69.0 & \cellcolor{tbblue!16}35.2 & \cellcolor{tbblue!8}53.0 & \cellcolor{tbblue!16}19.5 & \cellcolor{tbblue!8}43.0 & \cellcolor{tbblue!8}53.0 & \cellcolor{tbblue!16}27.2 & \cellcolor{tbblue!16}15.5 & \cellcolor{tbblue!16}0.322 & \cellcolor{tbblue!8}0.355 & \cellcolor{tbblue!16}23.2 \\
    & ReSum        & \cellcolor{tbred!30}44.5 & \cellcolor{tbred!30}21.5 & \cellcolor{tbred!24}40.5 & \cellcolor{tbred!18}15.8 & \cellcolor{tbred!18}37.5 & \cellcolor{tbred!24}43.5 & \cellcolor{tbred!18}23.2 & \cellcolor{tbblue!36}17.8 & \cellcolor{tbblue!16}0.324 & \cellcolor{tbblue!16}0.358 & \cellcolor{tbblue!16}23.8 \\
    & IterResearch & \cellcolor{tbred!18}62.3 & \cellcolor{tbred!18}29.5 & \cellcolor{tbred!18}44.5 & \cellcolor{tbred!18}16.5 & \cellcolor{tbred!30}24.5 & \cellcolor{tbred!30}36.5 & \cellcolor{tbred!30}15.0 & \cellcolor{tbred!30}7.5  & \cellcolor{tbblue!16}0.328 & \cellcolor{tbblue!16}0.362 & \cellcolor{tbblue!16}24.0 \\
    & U-Fold       & \cellcolor{tbblue!8}\textbf{70.2} & \cellcolor{tbblue!26}\underline{37.5} & \cellcolor{tbblue!16}\underline{54.5} & \cellcolor{tbblue!26}\underline{21.2} & \cellcolor{tbblue!26}\underline{48.0} & \cellcolor{tbblue!16}\textbf{55.8} & \cellcolor{tbblue!36}\underline{30.5} & \cellcolor{tbblue!46}\underline{19.8} & \cellcolor{tbblue!26}\underline{0.345} & \cellcolor{tbblue!26}\underline{0.380} & \cellcolor{tbblue!26}\underline{26.0} \\
    & IDSS & \cellcolor{tbblue!8}\underline{70.0} & \cellcolor{tbblue!36}\textbf{39.8} & \cellcolor{tbblue!26}\textbf{56.0} & \cellcolor{tbblue!46}\textbf{23.5} & \cellcolor{tbblue!36}\textbf{49.8} & \cellcolor{tbblue!16}\underline{55.5} & \cellcolor{tbblue!46}\textbf{32.2} & \cellcolor{tbblue!46}\textbf{22.0} & \cellcolor{tbblue!26}\textbf{0.365} & \cellcolor{tbblue!36}\textbf{0.398} & \cellcolor{tbblue!36}\textbf{28.2} \\
    \midrule
    \multirow{6}{*}{\makecell[l]{\textit{Claude-4.5-}\\\textit{Sonnet}}} & ReAct        & 67.0 & 32.5 & 51.5 & 17.8 & 43.5 & 53.0 & 27.5 & 15.0 & 0.318 & 0.350 & 24.0 \\
    & StateAct     & \cellcolor{tbblue!8}67.8 & \cellcolor{tbblue!16}33.8 & \cellcolor{tbblue!8}52.5 & \cellcolor{tbblue!16}18.8 & \cellcolor{tbblue!8}44.5 & \cellcolor{tbblue!8}54.2 & \cellcolor{tbblue!16}28.5 & \cellcolor{tbblue!26}16.2 & \cellcolor{tbblue!16}0.328 & \cellcolor{tbblue!8}0.360 & \cellcolor{tbblue!8}24.5 \\
    & ReSum        & \cellcolor{tbred!30}43.5 & \cellcolor{tbred!30}20.8 & \cellcolor{tbred!24}39.5 & \cellcolor{tbred!24}14.8 & \cellcolor{tbred!18}38.8 & \cellcolor{tbred!24}44.5 & \cellcolor{tbred!18}24.8 & \cellcolor{tbblue!36}18.3 & \cellcolor{tbblue!16}0.329 & \cellcolor{tbblue!16}0.362 & \cellcolor{tbblue!16}25.0 \\
    & IterResearch & \cellcolor{tbred!18}61.0 & \cellcolor{tbred!18}28.0 & \cellcolor{tbred!24}43.0 & \cellcolor{tbred!18}15.8 & \cellcolor{tbred!30}25.0 & \cellcolor{tbred!30}37.0 & \cellcolor{tbred!30}15.5 & \cellcolor{tbred!30}7.8  & \cellcolor{tbblue!16}0.334 & \cellcolor{tbblue!16}0.368 & \cellcolor{tbblue!16}25.5 \\
    & U-Fold       & \cellcolor{tbblue!16}\underline{69.5} & \cellcolor{tbblue!26}\underline{36.8} & \cellcolor{tbblue!16}\underline{54.0} & \cellcolor{tbblue!36}\underline{20.8} & \cellcolor{tbblue!26}\textbf{49.5} & \cellcolor{tbblue!16}\textbf{56.5} & \cellcolor{tbblue!36}\underline{31.8} & \cellcolor{tbblue!46}\underline{20.5} & \cellcolor{tbblue!26}\underline{0.350} & \cellcolor{tbblue!26}\underline{0.385} & \cellcolor{tbblue!26}\underline{27.0} \\
    & IDSS & \cellcolor{tbblue!16}\textbf{71.0} & \cellcolor{tbblue!36}\textbf{38.8} & \cellcolor{tbblue!26}\textbf{55.5} & \cellcolor{tbblue!46}\textbf{23.0} & \cellcolor{tbblue!26}\underline{49.2} & \cellcolor{tbblue!16}\underline{56.2} & \cellcolor{tbblue!46}\textbf{33.5} & \cellcolor{tbblue!46}\textbf{22.8} & \cellcolor{tbblue!36}\textbf{0.370} & \cellcolor{tbblue!36}\textbf{0.405} & \cellcolor{tbblue!36}\textbf{29.0} \\
    \midrule
    \multirow{6}{*}{\makecell[l]{\textit{Gemini-}\\\textit{2.5-Pro}}} & ReAct        & 69.0 & 34.5 & 53.0 & 19.0 & 44.0 & 54.5 & 27.8 & 15.8 & 0.320 & 0.352 & 23.0 \\
    & StateAct     & \cellcolor{tbblue!8}69.8 & \cellcolor{tbblue!16}35.8 & \cellcolor{tbblue!8}53.8 & \cellcolor{tbblue!16}20.0 & \cellcolor{tbblue!8}45.0 & \cellcolor{tbblue!8}55.2 & \cellcolor{tbblue!16}29.0 & \cellcolor{tbblue!16}16.8 & \cellcolor{tbblue!16}0.330 & \cellcolor{tbblue!8}0.362 & \cellcolor{tbblue!8}23.5 \\
    & ReSum        & \cellcolor{tbred!30}45.0 & \cellcolor{tbred!30}21.8 & \cellcolor{tbred!24}40.5 & \cellcolor{tbred!24}15.5 & \cellcolor{tbred!18}38.8 & \cellcolor{tbred!24}45.8 & \cellcolor{tbred!18}24.8 & \cellcolor{tbblue!36}19.5 & \cellcolor{tbblue!16}0.331 & \cellcolor{tbblue!16}0.364 & \cellcolor{tbblue!16}24.0 \\
    & IterResearch & \cellcolor{tbred!18}62.8 & \cellcolor{tbred!18}29.8 & \cellcolor{tbred!24}44.5 & \cellcolor{tbred!18}17.0 & \cellcolor{tbred!30}25.5 & \cellcolor{tbred!30}37.5 & \cellcolor{tbred!30}15.5 & \cellcolor{tbred!30}8.2  & \cellcolor{tbblue!16}0.338 & \cellcolor{tbblue!16}0.370 & \cellcolor{tbblue!16}24.5 \\
    & U-Fold       & \cellcolor{tbblue!8}\textbf{71.0} & \cellcolor{tbblue!26}\underline{38.0} & \cellcolor{tbblue!16}\underline{55.2} & \cellcolor{tbblue!26}\underline{21.8} & \cellcolor{tbblue!26}\underline{49.5} & \cellcolor{tbblue!16}\textbf{57.5} & \cellcolor{tbblue!26}\underline{31.5} & \cellcolor{tbblue!46}\underline{20.2} & \cellcolor{tbblue!26}\underline{0.348} & \cellcolor{tbblue!26}\underline{0.382} & \cellcolor{tbblue!26}\underline{26.2} \\
    & IDSS & \cellcolor{tbblue!8}\underline{70.5} & \cellcolor{tbblue!36}\textbf{40.2} & \cellcolor{tbblue!26}\textbf{57.0} & \cellcolor{tbblue!46}\textbf{24.0} & \cellcolor{tbblue!36}\textbf{50.0} & \cellcolor{tbblue!16}\underline{57.2} & \cellcolor{tbblue!46}\textbf{33.5} & \cellcolor{tbblue!46}\textbf{23.0} & \cellcolor{tbblue!36}\textbf{0.370} & \cellcolor{tbblue!36}\textbf{0.404} & \cellcolor{tbblue!36}\textbf{28.5} \\
    \midrule
    \multirow{6}{*}{\makecell[l]{\textit{DeepSeek-}\\\textit{V3.2}}} & ReAct        & 66.5 & 32.0 & 51.0 & 17.5 & 42.5 & 52.0 & 26.5 & 14.2 & 0.310 & 0.342 & 22.2 \\
    & StateAct     & \cellcolor{tbblue!8}67.2 & \cellcolor{tbblue!16}33.0 & \cellcolor{tbblue!8}51.8 & \cellcolor{tbblue!16}18.5 & \cellcolor{tbblue!8}43.5 & \cellcolor{tbblue!8}53.0 & \cellcolor{tbblue!16}27.5 & \cellcolor{tbblue!26}15.5 & \cellcolor{tbblue!16}0.320 & \cellcolor{tbblue!8}0.352 & \cellcolor{tbblue!8}23.0 \\
    & ReSum        & \cellcolor{tbred!30}43.2 & \cellcolor{tbred!30}20.2 & \cellcolor{tbred!24}39.0 & \cellcolor{tbred!24}14.8 & \cellcolor{tbred!18}37.5 & \cellcolor{tbred!24}43.8 & \cellcolor{tbred!18}23.5 & \cellcolor{tbblue!36}17.5 & \cellcolor{tbblue!16}0.321 & \cellcolor{tbblue!16}0.354 & \cellcolor{tbblue!16}23.2 \\
    & IterResearch & \cellcolor{tbred!18}60.5 & \cellcolor{tbred!18}27.5 & \cellcolor{tbred!24}42.8 & \cellcolor{tbred!18}15.5 & \cellcolor{tbred!30}24.5 & \cellcolor{tbred!30}36.0 & \cellcolor{tbred!30}15.0 & \cellcolor{tbred!30}7.5  & \cellcolor{tbblue!16}0.327 & \cellcolor{tbblue!16}0.359 & \cellcolor{tbblue!16}23.8 \\
    & U-Fold       & \cellcolor{tbblue!16}\underline{69.0} & \cellcolor{tbblue!26}\underline{36.0} & \cellcolor{tbblue!16}\underline{53.5} & \cellcolor{tbblue!36}\underline{20.5} & \cellcolor{tbblue!26}\textbf{48.2} & \cellcolor{tbblue!16}\textbf{55.8} & \cellcolor{tbblue!26}\underline{30.2} & \cellcolor{tbblue!46}\underline{19.5} & \cellcolor{tbblue!26}\underline{0.340} & \cellcolor{tbblue!26}\underline{0.374} & \cellcolor{tbblue!26}\underline{25.2} \\
    & IDSS & \cellcolor{tbblue!16}\textbf{69.5} & \cellcolor{tbblue!36}\textbf{38.0} & \cellcolor{tbblue!26}\textbf{55.0} & \cellcolor{tbblue!46}\textbf{22.5} & \cellcolor{tbblue!26}\underline{47.8} & \cellcolor{tbblue!16}\underline{55.5} & \cellcolor{tbblue!46}\textbf{32.0} & \cellcolor{tbblue!46}\textbf{21.5} & \cellcolor{tbblue!36}\textbf{0.358} & \cellcolor{tbblue!36}\textbf{0.392} & \cellcolor{tbblue!36}\textbf{27.8} \\
    \midrule
    \multirow{6}{*}{\makecell[l]{\textit{GPT-4.1-}\\\textit{mini}}} & ReAct        & 61.5 & 27.0 & 45.5 & 14.2 & 36.8 & 47.5 & 21.5 & 10.8 & 0.285 & 0.315 & 19.5 \\
    & StateAct     & \cellcolor{tbblue!8}62.0 & \cellcolor{tbblue!16}28.0 & \cellcolor{tbblue!8}46.0 & \cellcolor{tbblue!16}14.8 & \cellcolor{tbblue!8}37.5 & \cellcolor{tbblue!8}48.3 & \cellcolor{tbblue!16}22.4 & \cellcolor{tbblue!16}11.5 & \cellcolor{tbblue!8}0.292 & \cellcolor{tbblue!8}0.322 & \cellcolor{tbblue!16}20.2 \\
    & ReSum        & \cellcolor{tbred!30}40.0 & \cellcolor{tbred!30}17.0 & \cellcolor{tbred!24}34.5 & \cellcolor{tbred!24}11.8 & \cellcolor{tbred!18}32.5 & \cellcolor{tbred!24}40.0 & \cellcolor{tbred!18}19.0 & \cellcolor{tbblue!36}13.3 & \cellcolor{tbblue!16}0.295 & \cellcolor{tbblue!16}0.326 & \cellcolor{tbblue!16}20.8 \\
    & IterResearch & \cellcolor{tbred!18}56.0 & \cellcolor{tbred!18}23.2 & \cellcolor{tbred!24}37.8 & \cellcolor{tbred!18}12.8 & \cellcolor{tbred!30}21.5 & \cellcolor{tbred!30}32.8 & \cellcolor{tbred!30}12.0 & \cellcolor{tbred!30}5.6  & \cellcolor{tbblue!16}0.301 & \cellcolor{tbblue!16}0.331 & \cellcolor{tbblue!16}21.0 \\
    & U-Fold       & \cellcolor{tbblue!16}\underline{64.5} & \cellcolor{tbblue!36}\textbf{31.2} & \cellcolor{tbblue!16}\underline{48.5} & \cellcolor{tbblue!36}\underline{17.2} & \cellcolor{tbblue!26}\textbf{42.5} & \cellcolor{tbblue!26}\textbf{51.5} & \cellcolor{tbblue!36}\underline{26.0} & \cellcolor{tbblue!46}\underline{15.8} & \cellcolor{tbblue!26}\underline{0.318} & \cellcolor{tbblue!26}\underline{0.350} & \cellcolor{tbblue!36}\textbf{23.8} \\
    & IDSS & \cellcolor{tbblue!16}\textbf{64.8} & \cellcolor{tbblue!26}\underline{30.5} & \cellcolor{tbblue!26}\textbf{49.5} & \cellcolor{tbblue!46}\textbf{18.5} & \cellcolor{tbblue!26}\underline{42.0} & \cellcolor{tbblue!16}\underline{51.2} & \cellcolor{tbblue!46}\textbf{27.0} & \cellcolor{tbblue!46}\textbf{16.5} & \cellcolor{tbblue!26}\textbf{0.322} & \cellcolor{tbblue!26}\textbf{0.354} & \cellcolor{tbblue!36}\underline{23.2} \\
    \midrule
    \multirow{6}{*}{\makecell[l]{\textit{Gemini-}\\\textit{2.0-Flash}}} & ReAct        & 63.0 & 28.5 & 47.5 & 15.2 & 38.5 & 49.0 & 23.0 & 12.0 & 0.295 & 0.326 & 20.5 \\
    & StateAct     & \cellcolor{tbblue!8}63.8 & \cellcolor{tbblue!8}29.2 & \cellcolor{tbblue!8}48.0 & \cellcolor{tbblue!16}16.0 & \cellcolor{tbblue!8}39.2 & \cellcolor{tbblue!8}50.0 & \cellcolor{tbblue!16}24.0 & \cellcolor{tbblue!26}13.0 & \cellcolor{tbblue!16}0.304 & \cellcolor{tbblue!16}0.336 & \cellcolor{tbblue!16}21.2 \\
    & ReSum        & \cellcolor{tbred!30}41.0 & \cellcolor{tbred!30}18.0 & \cellcolor{tbred!24}36.0 & \cellcolor{tbred!24}12.8 & \cellcolor{tbred!18}34.0 & \cellcolor{tbred!24}41.2 & \cellcolor{tbred!18}20.5 & \cellcolor{tbblue!36}14.8 & \cellcolor{tbblue!16}0.305 & \cellcolor{tbblue!16}0.337 & \cellcolor{tbblue!16}21.5 \\
    & IterResearch & \cellcolor{tbred!18}57.5 & \cellcolor{tbred!18}24.5 & \cellcolor{tbred!24}39.5 & \cellcolor{tbred!18}13.5 & \cellcolor{tbred!30}22.5 & \cellcolor{tbred!30}33.8 & \cellcolor{tbred!30}13.0 & \cellcolor{tbred!30}6.2  & \cellcolor{tbblue!16}0.311 & \cellcolor{tbblue!16}0.342 & \cellcolor{tbblue!16}22.0 \\
    & U-Fold       & \cellcolor{tbblue!16}\underline{66.0} & \cellcolor{tbblue!26}\underline{32.5} & \cellcolor{tbblue!16}\underline{50.2} & \cellcolor{tbblue!36}\underline{18.2} & \cellcolor{tbblue!26}\textbf{43.5} & \cellcolor{tbblue!26}\textbf{53.0} & \cellcolor{tbblue!36}\underline{27.8} & \cellcolor{tbblue!46}\underline{17.0} & \cellcolor{tbblue!26}\underline{0.326} & \cellcolor{tbblue!26}\underline{0.358} & \cellcolor{tbblue!36}\textbf{24.5} \\
    & IDSS & \cellcolor{tbblue!16}\textbf{66.2} & \cellcolor{tbblue!36}\textbf{34.0} & \cellcolor{tbblue!26}\textbf{51.5} & \cellcolor{tbblue!46}\textbf{20.0} & \cellcolor{tbblue!26}\underline{43.2} & \cellcolor{tbblue!16}\underline{52.8} & \cellcolor{tbblue!36}\textbf{28.5} & \cellcolor{tbblue!46}\textbf{18.2} & \cellcolor{tbblue!26}\textbf{0.328} & \cellcolor{tbblue!26}\textbf{0.360} & \cellcolor{tbblue!36}\underline{24.2} \\
    \bottomrule
    \end{tabular}%
    }
    \end{table*}

\subsection{Experimental Setup}
\label{sec:setup}


We evaluate IDSS on three interactive benchmarks. \textbf{$\tau$-bench}~\citep{yao2024tau} includes Retail tasks on product inquiry, order management, and returns, and a more constraint-heavy Airline domain involving booking changes, cancellation, baggage, and fare-class restrictions. We report \textit{Avg@4} and \textit{pass\textsuperscript{4}} over four trials.
\textbf{VitaBench}~\citep{he2025vitabench} covers Delivery, In-store, OTA, and Cross-domain life-service tasks, where the latter two require more cross-service coordination and state inheritance, and reports rubric-based scores from 0 to 100.
\textbf{UserBench}~\citep{qian2025userbench} evaluates underspecified requests requiring preference elicitation, with \textit{Score}, \textit{CER} (Correct Exist Rate), and \textit{PE} (Preference Elicited) as metrics.

\begin{figure}[!t]
    \centering
    \includegraphics[width=0.95\linewidth]{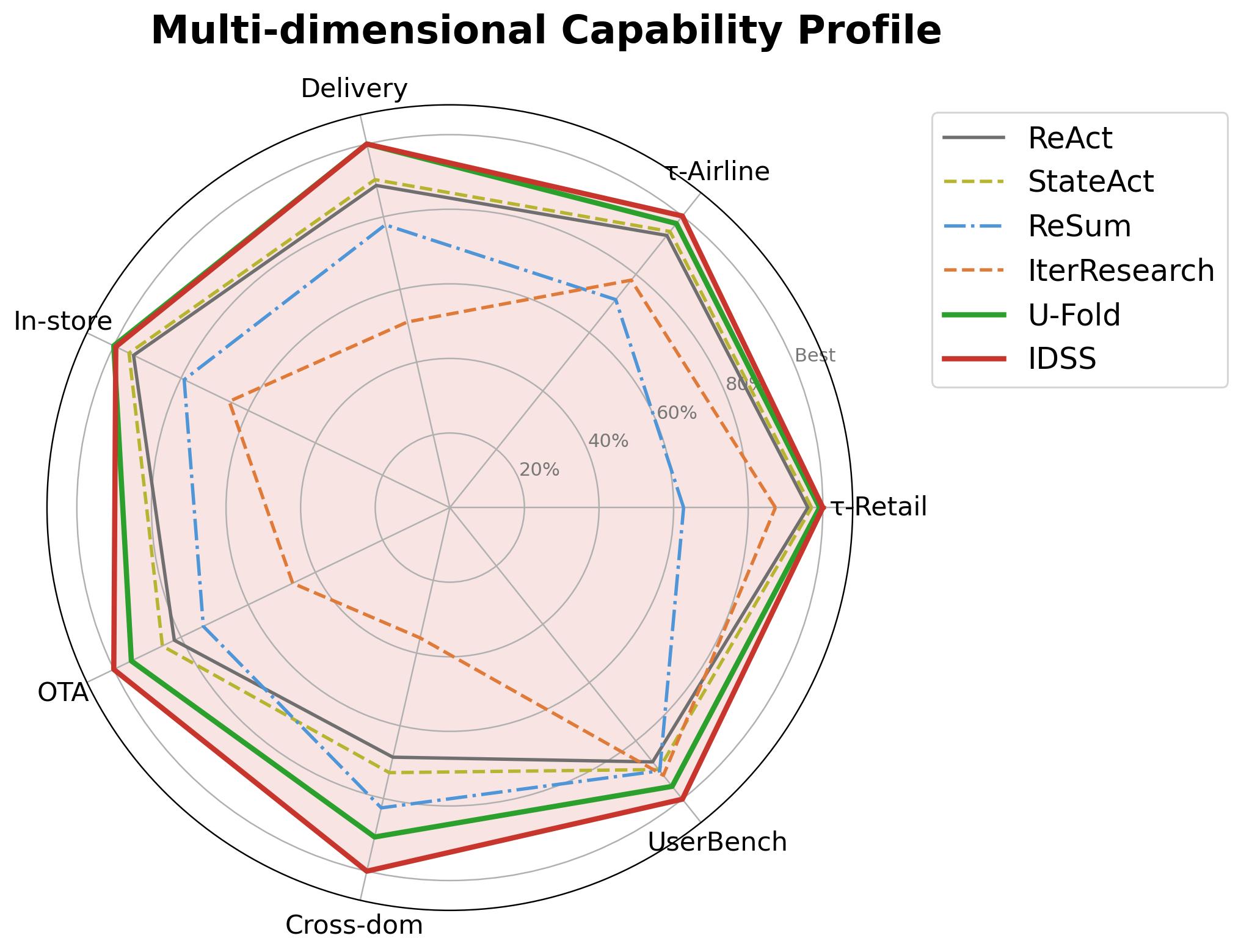}
    
    \caption{Normalized performance radar across seven evaluation dimensions. IDSS (red) achieves the largest coverage area.}
    \label{fig:radar}
\end{figure}

We compare against ReAct~\citep{yao2022react}, StateAct~\citep{rozanov2025stateact}, ReSum~\citep{wu2025resum}, IterResearch~\citep{chen2025iterresearch}, and U-Fold~\citep{su2026u} (see Appendix~\ref{sec:appendix_setup}). We use GPT-4.1 as the user simulator by following a common practice and evaluate on eight LLMs. For $\tau$-bench and VitaBench, each task is run for four independent trials and averaged.


\subsection{Main Results}
\label{sec:main_results}

Table~\ref{tab:main_results} reports the main results across three benchmarks and eight LLMs. Figure~\ref{fig:radar} summarizes the normalized performance profile across evaluation axes. Overall, IDSS delivers the strongest overall performance among the compared methods and consistently improves over ReAct across tool-use, life-service, and user-centric interaction settings. The radar plot further shows that IDSS has the most balanced coverage across evaluation axes, rather than improving only a single benchmark or metric. These results suggest that explicit situation states provide a general context-management benefit by helping agents preserve tool-grounded facts, track evolving user goals, and make constraint-aware decisions across different backbone models.

On $\tau$-bench, Airline is more challenging, as rebooking, cancellation, baggage handling, and fare-class restrictions introduce cross-turn dependencies and blocking conditions. Across the eight LLMs, IDSS improves over the strongest baseline, U-Fold, by 1.5 Avg@4 and 2.0 pass\textsuperscript{4} points on Airline. Retail gains over U-Fold are smaller but positive, at 0.6 Avg@4 and 1.8 pass\textsuperscript{4} points. In contrast, ReSum and IterResearch often underperform ReAct, indicating that summary-based or retrieval-heavy context management may lose precise identifiers, statuses, and time-sensitive fields required by customer-service tasks.

On VitaBench, U-Fold is the strongest baseline, especially on the simpler Delivery and In-store sub-tasks, where folding completed context already preserves much of the needed information. IDSS remains competitive on these simpler settings, while showing clearer advantages on OTA and Cross-domain, with gains over U-Fold of 1.7 and 2.1 points across the eight LLMs. These two scenarios require multi-entity coordination, cross-service composition, and state inheritance, where separating grounded facts from active intents and constraints provides more reliable guidance for subsequent actions.

On UserBench, IDSS achieves the best Score and CER across all evaluated LLMs, improving over U-Fold by approximately 4.6\% on Score and 4.0\% on CER in relative terms. Preference elicitation gains are smaller and vary across backbones, but the stronger completion-oriented metrics suggest that IDSS better distinguishes confirmed information, missing variables, and user preferences. This helps the agent ask clarification questions that are more directly tied to task progress, rather than eliciting preferences in isolation.

\subsection{Ablation Study}
\label{sec:ablation}

\begin{table}[!t]
    \centering
    \caption{Ablation results of IDSS on GPT-4.1 across three benchmarks.}
    \label{tab:ablation}
    \resizebox{0.95\columnwidth}{!}{%
    \begin{tabular}{l|cc|c|c}
    \toprule
    \textbf{Configuration} & \makecell{$\tau$-bench\\Avg@4} & \makecell{$\tau$-bench\\pass\textsuperscript{4}} & \makecell{VitaBench\\Avg@4} & \makecell{UserBench\\Score} \\
    \midrule
    Full IDSS     & \textbf{67.0} & \textbf{34.8} & \textbf{42.2} & \textbf{0.382} \\
    w/o Fact Layer        & 64.8 & 31.3 & 39.7 & 0.372 \\
    w/o State Layer       & 65.5 & 32.3 & 39.4 & 0.350 \\
    w/o Constraint Modeling  & 66.0 & 33.6 & 40.7 & 0.368 \\
    Vanilla               & 63.9 & 30.0 & 37.1 & 0.330 \\
    \bottomrule
    \end{tabular}%
    }
\end{table}

To understand where IDSS gains come from, we conduct ablation experiments on all three benchmarks by removing the fact layer, the state layer, or constraint modeling.  Specifically, \textit{w/o Fact Layer} removes tool-fact extraction and fact-layer rendering; \textit{w/o State Layer} removes intent tracking, variable tracking, and constraint modeling; and \textit{w/o Constraint Modeling} retains intent and variable tracking but removes action-precondition and blocking-reason modeling.


The results reveal complementary contributions from the fact and state layers. Removing the fact layer hurts $\tau$-bench most, reducing Avg@4 by 2.2 points and pass\textsuperscript{4} by 3.5 points, because these tasks depend heavily on persistent tool-grounded facts such as orders, fares, and reservations. Removing the state layer has the largest impact on UserBench, dropping Score by 8.4\%, versus 2.6\% without the fact layer, reflecting the importance of intent progress and missing-variable tracking for preference elicitation. On VitaBench, the two layers contribute similarly, with drops of 2.5 and 2.8 points, indicating that complex life-service tasks require both factual grounding and evolving user-goal tracking. Constraint modeling provides smaller but consistent gains by translating factual changes into executable or blocked intents.




\subsection{Error Analysis}
\label{sec:error}

\begin{figure}[!t]
    \centering
    \includegraphics[width=\linewidth]{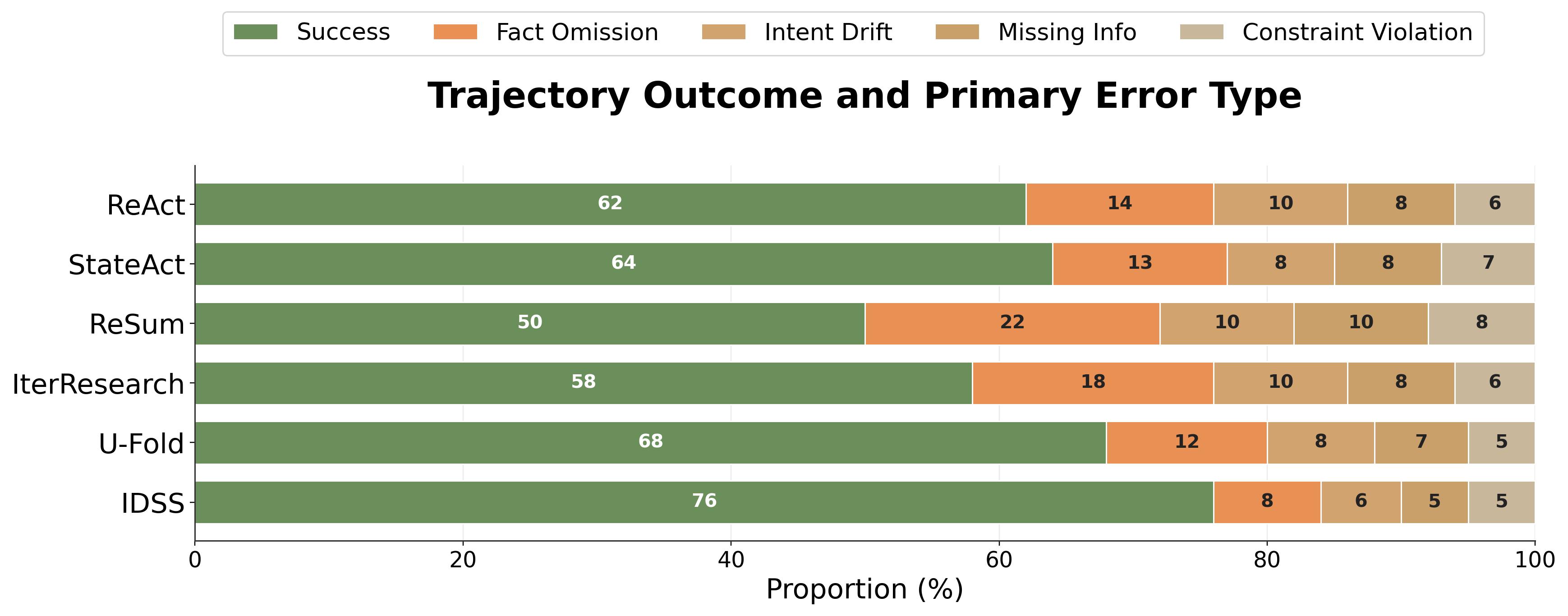}
    \caption{Trajectory outcomes and primary error types for IDSS and baselines.}
    \label{fig:error}
\end{figure}

To better understand the error patterns behind the aggregate results, we annotate $\tau$-bench and VitaBench trajectories into the outcome and error categories shown in Figure~\ref{fig:error}. IDSS substantially reduces errors related to state maintenance. In particular, fact omission decreases from 14\% to 8\%, reflecting the benefit of persistent entity storage in the fact layer. Missing-information errors also decrease from 8\% to 5\%, consistent with the state layer's explicit tracking of missing variables.

Different baselines exhibit distinct failure patterns. ReSum has the highest fact-omission rate at 22\%, suggesting that summarization can discard fine-grained fields needed for later decisions. StateAct reduces intent drift, but its fact-omission rate remains close to ReAct because it does not explicitly preserve tool-grounded attributes. In contrast, constraint violation changes only modestly across methods, ranging from 5\% to 8\%. This suggests that IDSS is most effective at mitigating state-maintenance errors, while some rule violations still arise from the model's imperfect rule understanding rather than from missing state alone.


\subsection{Efficiency Analysis}
\label{sec:efficiency}

\begin{figure*}[!t]
    \centering
    \begin{subfigure}[t]{0.56\textwidth}
        \centering
        \includegraphics[width=\linewidth]{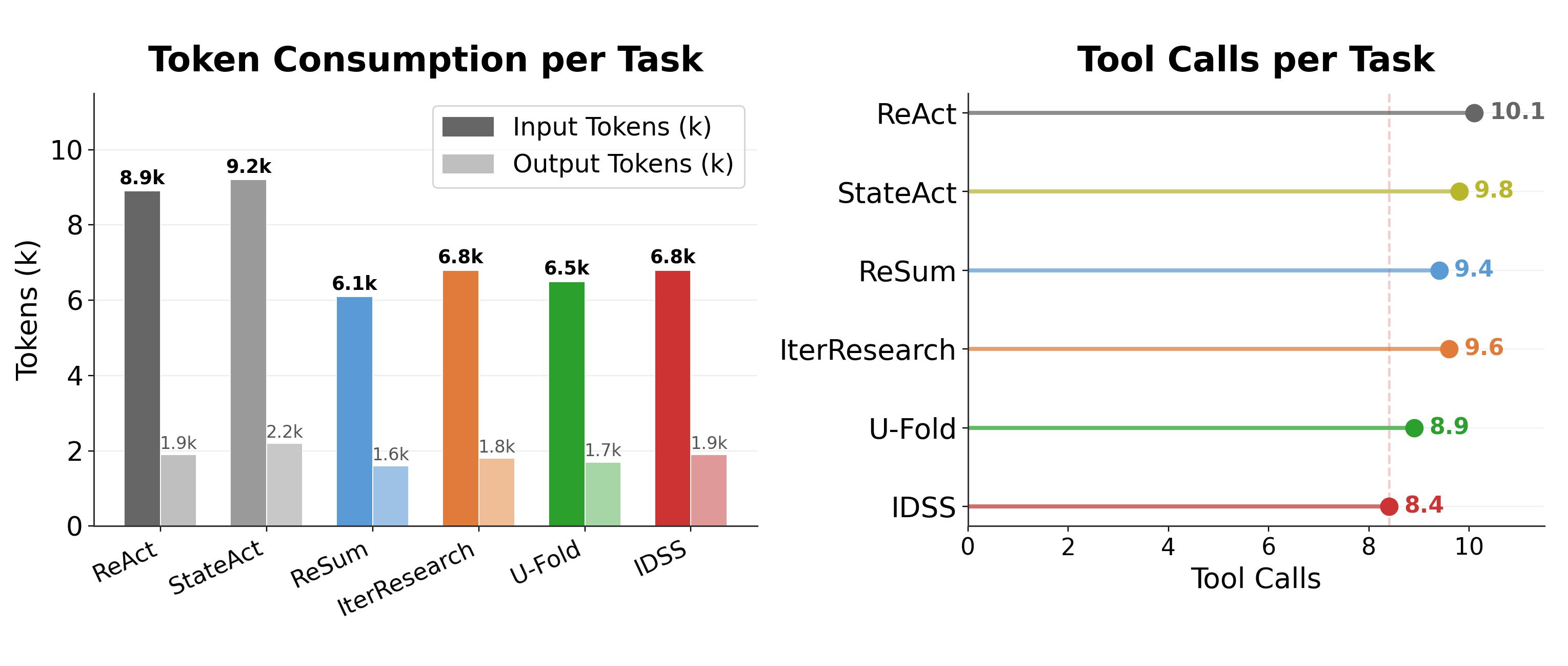}
        \caption{Average input/output token usage and tool calls per task.}
        \label{fig:token}
    \end{subfigure}
    \hfill
    \begin{subfigure}[t]{0.38\textwidth}
        \centering
        \includegraphics[width=\linewidth]{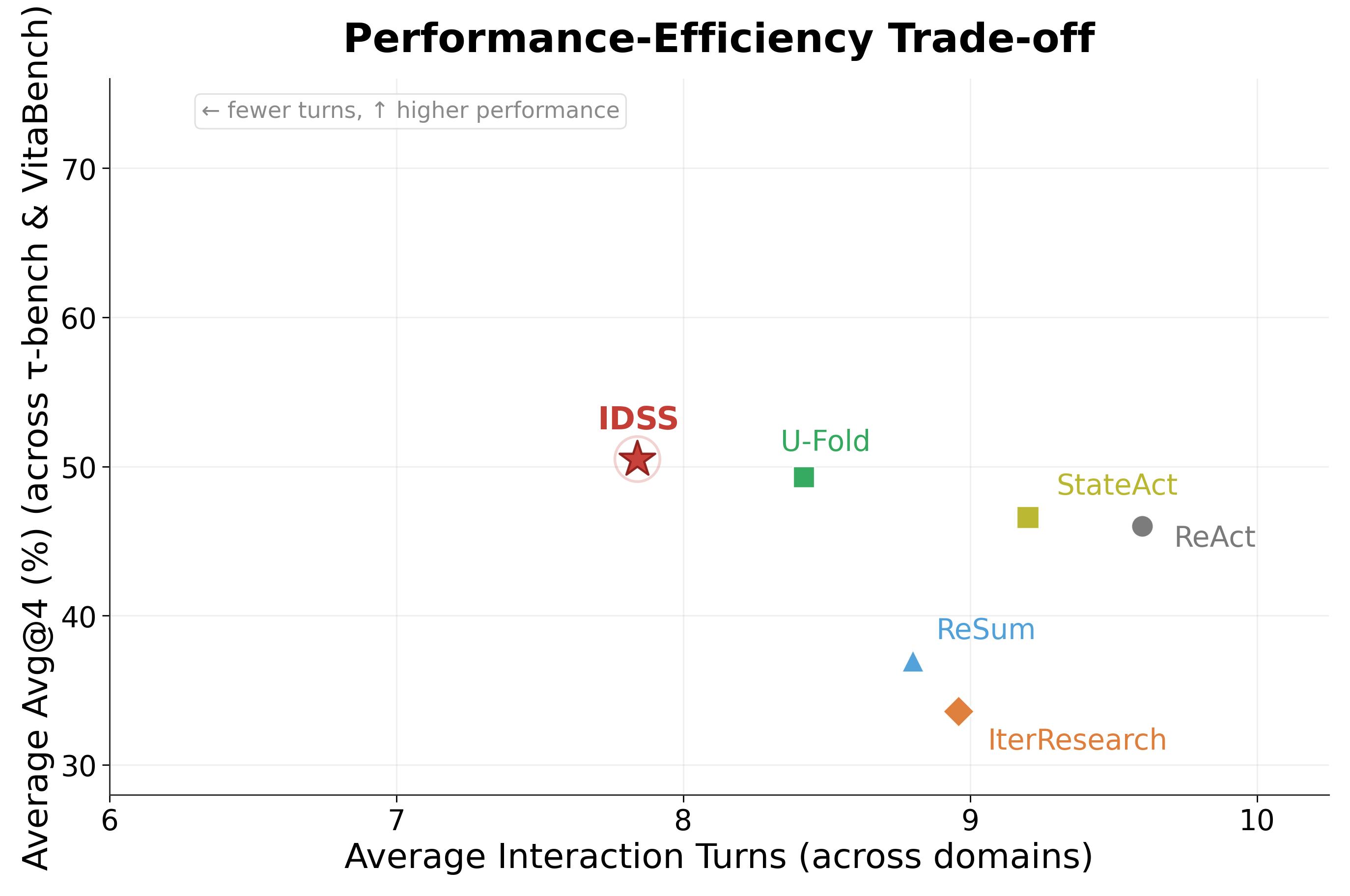}
        \caption{Performance versus interaction turns.}
        \label{fig:pareto}
    \end{subfigure}
    \caption{Efficiency comparison of IDSS and baselines.}
    \label{fig:efficiency}
\end{figure*}

\begin{figure*}[!t]
    \centering
    \includegraphics[width=\textwidth]{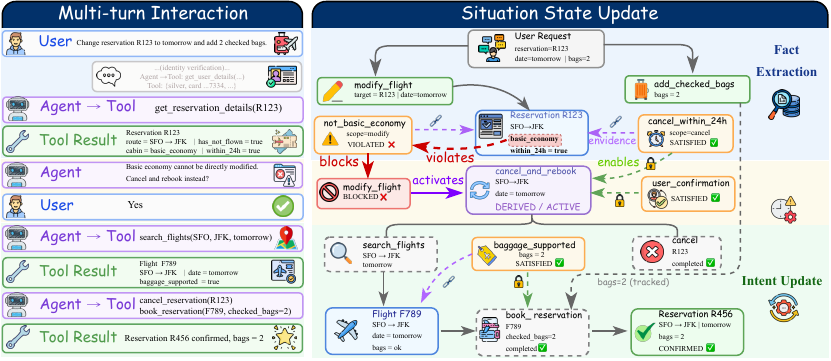}
    \caption{Case study on a $\tau$-bench Airline task. IDSS propagates a tool-grounded fare-class fact to block an infeasible modification path and activate a feasible alternative while preserving the user's baggage requirement.}
    \label{fig:case}
\end{figure*}

Figure~\ref{fig:token} compares token consumption and tool-call counts, while Figure~\ref{fig:pareto} plots average performance against interaction turns. IDSS keeps the prompt compact by rendering the current situation state with compressed history, requires \textit{zero} extra LLM calls, and achieves the highest average performance (50.5\%) with the fewest turns (7.8). These gains come from preserving reusable facts and preventing actions on blocked or underspecified intents.

\subsection{Case Study}
\label{sec:case_study}

Figure~\ref{fig:case} presents a $\tau$-bench Airline case where a \texttt{basic\_economy} fare-class fact must be linked to the modification constraint. History-only or summary-based agents may preserve the user's request but miss the blocking relation between \texttt{basic\_economy} and \texttt{modify\_flight}. IDSS records this fact during extraction, blocks the infeasible \texttt{modify\_flight} intent through constraint propagation, and activates an alternative path while preserving the checked bag requirement.

\section{Conclusion}
\label{sec:conclusion}

In this work, we presented IDSS, a training-free framework for intent-driven situation tracking in user-centric multi-turn agents.
IDSS maintains the current task situation as an explicit state, separating tool-grounded facts from evolving task-state judgments such as intents, missing variables, constraints, and execution status.
Across three interactive benchmarks and eight LLMs, IDSS improves task completion, preference elicitation, and interaction efficiency, especially on tasks with multi-entity coordination and evolving constraints.
These results suggest that reliable user-centric agents benefit from maintaining an explicit, decision-oriented situation state rather than relying only on access to past interaction history.

\section*{Limitations}
\label{sec:limitations}

IDSS is designed for interactive tasks where tool returns expose textual fields that can be converted into structured facts. Extending the same situation-state design to more diverse output formats, richer tool environments, or multimodal observations is an interesting direction for future work. Our experiments use simulated users for reproducibility and controlled comparison across models. Future studies with real users could further validate how IDSS supports naturally expressed goals, corrections, and preferences.

\section*{Ethical Considerations}

This work studies training-free context management for user-centric multi-turn agents using public or simulated benchmark environments. We do not collect personal data or conduct experiments with human participants. IDSS may improve the reliability of tool-using agents, but it does not eliminate risks from incorrect tool outputs, model hallucinations, or inappropriate deployment in high-stakes settings. Practical deployments should include task-specific safeguards, logging, and human oversight when agent actions may affect users.

\bibliography{reference}

\clearpage
\appendix
\raggedbottom
\emergencystretch=1em

\section{Experimental Setup Details}
\label{sec:appendix_setup}

\subsection{Benchmarks}

\paragraph{$\tau$-bench.}
$\tau$-bench~\citep{yao2024tau} simulates realistic multi-turn dialogues between an LLM agent and a user simulator in domain-specific environments equipped with API tools and rules. It provides two domains: \textbf{Retail} (product inquiry, order management, returns) and \textbf{Airline} (flight booking, modification, cancellation with fare-class restrictions). Each task involves 3 to 8 interaction turns with the user simulator issuing follow-up requests conditioned on prior agent responses. Evaluation uses reward-based metrics: \textit{Avg@k} measures the mean task-completion reward over $k$ independent trials, and \textit{pass\textsuperscript{k}} measures the probability that all $k$ trials succeed.

\paragraph{VitaBench.}
VitaBench~\citep{he2025vitabench} is a life-service simulation benchmark requiring agents to complete complex, tool-intensive tasks across four sub-scenarios: \textbf{Delivery} (food ordering with address and time constraints), \textbf{In-store} (restaurant reservation and service inquiries), \textbf{OTA} (travel planning involving transportation, accommodation, and attraction coordination), and \textbf{Cross-domain} (tasks spanning multiple service categories with shared state). Tasks feature verbose tool outputs, multi-entity coordination, and progressively revealed user constraints. Evaluation uses rubric-based scoring from 0 to 100 assessed by a GPT-4.1~\citep{openai2025gpt41} judge.

\paragraph{UserBench.}
UserBench~\citep{qian2025userbench} evaluates agents on realistic user interactions where goals are underspecified and incrementally revealed. Unlike benchmarks with fully specified task descriptions, UserBench requires agents to proactively clarify ambiguous requirements and elicit user preferences. It reports three metrics: \textit{Score} (overall task completion quality), \textit{CER} (Correct Exist Rate, measuring whether the agent correctly identifies existing information), and \textit{PE} (Preference Elicited, measuring the agent's ability to actively elicit user preferences before acting).

\subsection{Baselines}

\begin{itemize}
    \item \textbf{ReAct}~\citep{yao2022react} retains complete reasoning-action-observation traces without compression, serving as the standard agentic paradigm.
    \item \textbf{StateAct}~\citep{rozanov2025stateact} augments ReAct with chain-of-states, explicitly generating goal, state, thought, and action at each turn, but does not separate tool-grounded facts from task-state judgments.
    \item \textbf{ReSum}~\citep{wu2025resum} periodically invokes an external summarizer to compress interaction history into compact reasoning states, controlling context growth at the cost of additional LLM calls.
    \item \textbf{IterResearch}~\citep{chen2025iterresearch} restructures the workspace into an evolving report, synthesizing new findings at each turn while discarding raw interaction traces.
    \item \textbf{U-Fold}~\citep{su2026u} dynamically generates intent-aware dialogue summaries and extracts compact, task-relevant tool logs at each turn through two LLM-based modules.
\end{itemize}

\section{Prompt Templates}
\label{sec:appendix_prompts}

\subsection{System Prompt with IDSS State}

\begin{tcolorbox}[
    colback=promptbg,
    colframe=promptframe,
    title={\textbf{System Prompt with IDSS State}},
    fonttitle=\small\sffamily,
    coltitle=white,
    colbacktitle=promptframe,
    boxrule=0.5pt,
    arc=2pt,
    left=6pt, right=6pt, top=4pt, bottom=4pt,
    fontupper=\small\ttfamily,
    breakable,
    pad at break=3pt
]
You are a task-oriented assistant. You help users complete multi-step tasks by calling tools, gathering information, and taking actions on behalf of the user.\\[4pt]
\textcolor{promptframe}{\rule{\linewidth}{0.4pt}}\\[2pt]
\textcolor{promptframe}{\textbf{BEHAVIORAL GUIDELINES}}\\[2pt]
- Always verify facts from the Fact Layer before making claims. Never fabricate entity attributes.\\
- If a required variable is unknown, ask the user (if askable) or call a tool (if retrievable).\\
- Before executing an action, check all associated constraints. Do not proceed if any constraint is in VIOLATED status.\\
- When an intent is BLOCKED, explain the reason to\\
\hspace*{1em}the user and propose feasible alternatives.\\
- After completing a sub-task, review remaining PENDING intents before ending the conversation.\\[4pt]
\textcolor{promptframe}{\rule{\linewidth}{0.4pt}}\\[2pt]
\textcolor{promptframe}{\textbf{CURRENT TASK STATE (IDSS)}}\\[2pt]
The following is your structured understanding of the current task. Use it as your primary decision-making reference. The state is automatically maintained across turns. You must update it after each response.\\[4pt]
\textcolor{promptframe}{\textbf{=== FACT LAYER (verified entities) ===}}\\
\{fact\_layer\_snapshot\}\\[4pt]
\textcolor{promptframe}{\textbf{=== STATE LAYER (task understanding) ===}}\\
\{state\_layer\_snapshot\}\\[4pt]
\textcolor{promptframe}{\rule{\linewidth}{0.4pt}}\\[2pt]
\textcolor{promptframe}{\textbf{STATE MAINTENANCE PROTOCOL}}\\[2pt]
For EVERY response, update the <state> tag with the current state layer. Follow this protocol:\\[2pt]
1. \textcolor{stateframe}{\textbf{Intents}}: Update status of each intent:\\
\hspace*{1em}- ACTIVE $\rightarrow$ currently processing\\
\hspace*{1em}- PENDING $\rightarrow$ known but not yet started\\
\hspace*{1em}- COMPLETED $\rightarrow$ finished (record outcome)\\
\hspace*{1em}- BLOCKED $\rightarrow$ infeasible (record reason)\\
\hspace*{1em}Add new intents when the user introduces new goals.\\[2pt]
2. \textcolor{stateframe}{\textbf{Variables}}: For each information slot:\\
\hspace*{1em}- Mark source: ASKABLE / RETRIEVABLE / DERIVABLE\\
\hspace*{1em}- Update value and known status after each turn.\\
\hspace*{1em}- Add new variables as task requirements emerge.\\[2pt]
3. \textcolor{stateframe}{\textbf{Constraints}}: For each constraint:\\
\hspace*{1em}- SATISFIED $\rightarrow$ condition currently met\\
\hspace*{1em}- UNSATISFIED $\rightarrow$ not yet verifiable\\
\hspace*{1em}- VIOLATED $\rightarrow$ confirmed impossible\\
\hspace*{1em}When VIOLATED, mark the associated intent\\
\hspace*{1em}as BLOCKED with factual evidence.
\end{tcolorbox}

\subsection{State Tag Output Format}

Below is a concrete example from a $\tau$-bench airline scenario where the user requests a flight modification and two checked bags.

\begin{tcolorbox}[
    colback=statebg,
    colframe=stateframe,
    title={\textbf{State Tag Example ($\tau$-bench Airline Scenario)}},
    fonttitle=\small\sffamily,
    coltitle=white,
    colbacktitle=stateframe,
    boxrule=0.5pt,
    arc=2pt,
    left=6pt, right=6pt, top=4pt, bottom=4pt,
    fontupper=\small\ttfamily,
    breakable,
    pad at break=3pt
]
<state>\\
\textcolor{stateframe}{\textbf{intents:}}\\
- name: "modify\_flight"\\
\hspace*{1em}status: blocked\\
\hspace*{1em}reason: "R123.cabin = basic\_economy\\
\hspace*{3em}violates not\_basic\_economy"\\
\hspace*{1em}depends\_on: []\\
- name: "cancel\_and\_rebook"\\
\hspace*{1em}status: active\\
\hspace*{1em}reason: "alternative enabled by blocked\\
\hspace*{3em}modification + cancel\_within\_24h"\\
\hspace*{1em}depends\_on: ["modify\_flight.blocked"]\\
- name: "add\_checked\_bags"\\
\hspace*{1em}status: pending\\
\hspace*{1em}depends\_on:\\
\hspace*{2em}["cancel\_and\_rebook.completed"]\\[4pt]
\textcolor{stateframe}{\textbf{variables:}}\\
- name: "reservation\_id"\\
\hspace*{1em}value: "R123"\\
\hspace*{1em}source: askable | known: true\\
- name: "original\_route"\\
\hspace*{1em}value: "SFO to JFK"\\
\hspace*{1em}source: retrievable | known: true\\
\hspace*{1em}evidence: "Reservation R123.route"\\
- name: "checked\_bags"\\
\hspace*{1em}value: 2\\
\hspace*{1em}source: askable | known: true\\
- name: "replacement\_flight"\\
\hspace*{1em}value: null\\
\hspace*{1em}source: retrievable | known: false\\[4pt]
\textcolor{stateframe}{\textbf{constraints:}}\\
- scope: "flight\_modification"\\
\hspace*{1em}condition: "fare is not basic economy"\\
\hspace*{1em}status: violated\\
\hspace*{1em}evidence: "R123.cabin = basic\_economy"\\
- scope: "alternative\_booking"\\
\hspace*{1em}condition: "cancel within 24h, user confirmed"\\
\hspace*{1em}status: satisfied\\
\hspace*{1em}evidence: "within\_24h = true,\\
\hspace*{3em}user\_confirmation = yes"\\
- scope: "new\_reservation"\\
\hspace*{1em}condition: "replacement supports\\
\hspace*{3em}checked bags"\\
\hspace*{1em}status: unsatisfied\\
\hspace*{1em}evidence: null\\
</state>
\end{tcolorbox}

\subsection{Fact Layer Entity Format}

Below is an example from a $\tau$-bench airline task.

\begin{tcolorbox}[
    colback=factbg,
    colframe=factframe,
    title={\textbf{Fact Layer Snapshot ($\tau$-bench Airline Task)}},
    fonttitle=\small\sffamily,
    coltitle=white,
    colbacktitle=factframe,
    boxrule=0.5pt,
    arc=2pt,
    left=6pt, right=6pt, top=4pt, bottom=4pt,
    fontupper=\small\ttfamily,
    breakable,
    pad at break=3pt
]
\textcolor{factframe}{\textbf{[Entity: Reservation \#RSV-20240315]}}\\
\hspace*{1em}type: reservation\\
\hspace*{1em}id: RSV-20240315\\
\hspace*{1em}source: get\_reservation("RSV-20240315")\\[2pt]
\hspace*{1em}passenger\_name: "Alice Chen"\\
\hspace*{1em}flight\_number: UA-1234\\
\hspace*{1em}date: 2024-03-20\\
\hspace*{1em}departure: SFO 09:30 $\rightarrow$ JFK 18:05\\
\hspace*{1em}cabin\_class: economy\\
\hspace*{1em}booking\_status: confirmed\\
\hspace*{1em}change\_fee: \$150\\
\hspace*{1em}payment\_method: Visa ending 4392\\
\hspace*{1em}baggage\_allowance: 1 x 23kg\\
\hspace*{1em}seat\_assignment: 24A (window)\\[4pt]
\textcolor{factframe}{\textbf{[Entity: Flight UA-5678]}}\\
\hspace*{1em}type: flight (search result)\\
\hspace*{1em}id: UA-5678\\
\hspace*{1em}source: search\_flights(\{origin: "SFO", dest: "JFK",\\
\hspace*{4em}date: "2024-03-20", cabin: "economy"\})\\[2pt]
\hspace*{1em}route: SFO 07:15 $\rightarrow$ JFK 15:50\\
\hspace*{1em}cabin\_class: economy\\
\hspace*{1em}seats\_available: 12\\
\hspace*{1em}price\_difference: +\$45\\
\hspace*{1em}change\_eligible: true\\[4pt]
\textcolor{factframe}{\textbf{[Entity: Flight UA-9012]}}\\
\hspace*{1em}type: flight (search result)\\
\hspace*{1em}id: UA-9012\\
\hspace*{1em}source: search\_flights(\{...\})\\[2pt]
\hspace*{1em}route: SFO 11:30 $\rightarrow$ JFK 20:15\\
\hspace*{1em}cabin\_class: economy\\
\hspace*{1em}seats\_available: 3\\
\hspace*{1em}price\_difference: +\$20\\
\hspace*{1em}change\_eligible: true
\end{tcolorbox}


\section{Examples}
\label{sec:appendix_examples}

\subsection{\texorpdfstring{$\tau$-bench}{tau-bench} Retail: Multi-Item Exchange with Conditional Preferences}

\begin{tcolorbox}[
    colback=white,
    colframe=promptframe,
    title={\textbf{Example 1: Multi-item exchange with conditional fallback preferences}},
    fonttitle=\small\sffamily\bfseries,
    coltitle=white,
    colbacktitle=promptframe,
    boxrule=0.5pt, arc=3pt,
    left=8pt, right=8pt, top=6pt, bottom=6pt,
    fontupper=\small,
    breakable,
    pad at break=3pt
]

\noindent\textbf{Benchmark and task}: $\tau$-bench Retail, multi-item exchange with a conditional fallback preference.\\[2pt]
\textbf{User goal}:\\[2pt]
\textit{``You are Yusuf Rossi in 19122. You received order \#W2378156 and wish to exchange the mechanical keyboard for one with clicky switches, and the smart thermostat for one compatible with Google Home instead of Apple HomeKit. If there is no clicky + RGB + full size keyboard, you'd go for no backlight.''}

\tcbline

\textbf{Turn 1} \textcolor{promptframe}{\textbf{[User]}}\\
\textit{``Hi, I received order \#W2378156. I want to exchange two items: the mechanical keyboard, for which I'd like one with clicky switches, RGB backlight, full size. And the smart thermostat, for which I need Google Home compatible instead of Apple HomeKit.''}

\tcbline

\textbf{Turn 2} \textcolor{factframe}{\textbf{[Tool Call]}} \texttt{find\_user\_id\_by\_name\_zip}\\[2pt]
\begin{tcolorbox}[colback=factbg, colframe=factframe!30, boxrule=0.3pt, arc=1pt, left=4pt, right=4pt, top=2pt, bottom=2pt, fontupper=\small\ttfamily]
\{~first\_name: "Yusuf", last\_name: "Rossi", zip: "19122"~\}\\
$\rightarrow$ "yusuf\_rossi\_9620"
\end{tcolorbox}

\textbf{Turn 3} \textcolor{factframe}{\textbf{[Tool Call]}}\\
\texttt{get\_order\_details("\#W2378156")}\\[2pt]
\begin{tcolorbox}[colback=factbg, colframe=factframe!30, boxrule=0.3pt, arc=1pt, left=4pt, right=4pt, top=2pt, bottom=2pt, fontupper=\small\ttfamily]
\{\\
~~order\_id: "\#W2378156",\\
~~status: "delivered",\\
~~items: [\\
~~~~\{ name: "Headphones",~~item\_id: "4202497723",\\
~~~~~~price: 342.81, options: \{type: "over-ear", connectivity: "wireless"\} \},\\
~~~~\{ name: "Vacuum Cleaner", item\_id: "4602305039", price: 561.05 \},\\
~~~~\textbf{\{ name: "Mechanical Keyboard", item\_id: "1151293680",}\\
~~~~~~\textbf{product\_id: "1656367028",}\\
~~~~~~\textbf{price: 272.33,}\\
~~~~~~\textbf{options: \{switch\_type: "linear", backlight: "RGB", size: "full size"\} \},}\\
~~~~\textbf{\{ name: "Smart Thermostat", item\_id: "4983901480",}\\
~~~~~~\textbf{product\_id: "4896585277",}\\
~~~~~~\textbf{price: 262.47,}\\
~~~~~~\textbf{options: \{compatibility: "Apple HomeKit", color: "black"\} \},}\\
~~~~\{ name: "Smart Watch", item\_id: "9408160950", price: 381.26 \}\\
~~],\\
~~payment\_method: "credit\_card\_9513926"\\
\}
\end{tcolorbox}

\textbf{Turn 4} \textcolor{factframe}{\textbf{[Tool Call]}}\\
\texttt{get\_product\_details("1656367028")}\\[2pt]
\begin{tcolorbox}[colback=factbg, colframe=factframe!30, boxrule=0.3pt, arc=1pt, left=4pt, right=4pt, top=2pt, bottom=2pt, fontupper=\small\ttfamily]
\{ name: "Mechanical Keyboard", variants: [\\
~~~~\{ item\_id: "9690244451",\\
~~~~~~options: \{switch: "clicky", backlight: "RGB", size: "60\%"\},\\
~~~~~~\textbf{available: false} \},\\[2pt]
~~~~\textbf{\{ item\_id: "7706410293",}\\
~~~~~~\textbf{options: \{switch: "clicky", backlight: "none", size: "full size"\},}\\
~~~~~~\textbf{available: true, price: 269.16 \},}\\[2pt]
~~~~\{ item\_id: "6342039236",\\
~~~~~~options: \{switch: "clicky", backlight: "white", size: "full size"\},\\
~~~~~~available: true, price: 244.91 \},\\
~~~~...\\
] \}
\end{tcolorbox}

\textbf{Turn 5} \textcolor{factframe}{\textbf{[Tool Call]}}\\
\texttt{get\_product\_details("4896585277")}\\[2pt]
\begin{tcolorbox}[colback=factbg, colframe=factframe!30, boxrule=0.3pt, arc=1pt, left=4pt, right=4pt, top=2pt, bottom=2pt, fontupper=\small\ttfamily]
\{ name: "Smart Thermostat", variants: [\\
~~~~...\\
~~~~\textbf{\{ item\_id: "7747408585",}\\
~~~~~~\textbf{options: \{compatibility: "Google Assistant", color: "black"\},}\\
~~~~~~\textbf{available: true, price: 249.01 \},}\\[2pt]
~~~~\{ item\_id: "8722653925",\\
~~~~~~options: \{compatibility: "Google Assistant", color: "white"\},\\
~~~~~~available: false \},\\
~~~~...\\
] \}
\end{tcolorbox}

\begin{tcolorbox}[colback=statebg, colframe=stateframe, boxrule=0.4pt, arc=2pt, left=4pt, right=4pt, top=3pt, bottom=3pt, fontupper=\footnotesize\ttfamily]
\textcolor{stateframe}{\textbf{IDSS State (After Turn 5)}}\\[2pt]
\textcolor{stateframe}{\textbf{Fact Layer:}}\\
~~[Entity: Order] id: \#W2378156\\
~~~~source: get\_order\_details()\\
~~~~status: delivered, payment: "credit\_card\_9513926"\\
~~[Entity: Product] id: 1656367028 (Keyboard)\\
~~~~source: get\_product\_details()\\
~~~~variants: [9690244451 (clicky+RGB+60\%, \textbf{unavail.}),\\
~~~~~~~~~~~~~~7706410293 (clicky+none+full, \$269.16)]\\
~~[Entity: Product] id: 4896585277 (Thermostat)\\
~~~~source: get\_product\_details()\\
~~~~variants: [7747408585 (Google+black, \$249.01)]\\[2pt]
\textcolor{stateframe}{\textbf{State Layer:}}\\[1pt]
~~\textcolor{stateframe}{\textit{Intents:}}\\
~~~~I1: "Exchange keyboard" $\rightarrow$ \textbf{blocked}\\
~~~~~~\textit{reason}: primary variant unavailable\\
~~~~I1-fallback: "No backlight" $\rightarrow$ \textbf{active}\\
~~~~~~\textit{depends\_on}: [I1.blocked]\\
~~~~I2: "Exchange thermostat" $\rightarrow$ \textbf{active}\\[1pt]
~~\textcolor{stateframe}{\textit{Variables:}}\\
\hspace*{1em}\textbf{payment\_method}: "credit\_card\_9513926"\\
\hspace*{2em}source: derivable | known: true\\
\hspace*{2em}evidence: "order \#W2378156.payment"\\
\hspace*{1em}\textbf{price\_diff\_keyboard}: -\$3.17\\
\hspace*{2em}source: derivable | known: true\\
\hspace*{2em}evidence: "269.16 $-$ 272.33"\\
\hspace*{1em}\textbf{price\_diff\_thermostat}: -\$13.46\\
\hspace*{2em}source: derivable | known: true\\
\hspace*{2em}evidence: "249.01 $-$ 262.47"\\[1pt]
~~\textcolor{stateframe}{\textit{Constraints:}}\\
\hspace*{1em}C1: "clicky+RGB+full available" $\rightarrow$ \textbf{violated}\\
\hspace*{2em}evidence: "9690244451 unavailable"\\
\hspace*{1em}C2: "fallback: no backlight" $\rightarrow$ \textbf{satisfied}\\
\hspace*{2em}evidence: "C1 violated, user fallback"\\
\hspace*{1em}C3: "Google Home variant" $\rightarrow$ \textbf{satisfied}\\
\hspace*{2em}evidence: "7747408585 available"\\[1pt]
~~\textcolor{stateframe}{\textit{Next Action:}}\\
\hspace*{1em}Confirm both exchanges before calling API.
\end{tcolorbox}

\tcbline

\textbf{Turn 6} \textcolor{stateframe}{\textbf{[IDSS Agent]}}\\
\textit{``I checked the options. The clicky + RGB + full size keyboard is out of stock. Per your fallback preference, I'll use the clicky + no backlight + full size variant (\$269.16).}\\
\textit{For the thermostat, the Google Assistant model in black is available (\$249.01). Shall I proceed with both exchanges?''}

\tcbline

\textbf{Turn 7} \textcolor{promptframe}{\textbf{[User]}} \textit{``Yes, go ahead.''}

\tcbline

\textbf{Turn 8} \textcolor{factframe}{\textbf{[Tool Call]}}\\
\texttt{exchange\_delivered\_order\_items}\\[2pt]
\begin{tcolorbox}[colback=factbg, colframe=factframe!30, boxrule=0.3pt, arc=1pt, left=4pt, right=4pt, top=2pt, bottom=2pt, fontupper=\small\ttfamily]
\{\\
~~order\_id: "\#W2378156",\\
~~item\_ids: ["1151293680", "4983901480"],\\
~~new\_item\_ids: ["7706410293", "7747408585"],\\
~~payment\_method\_id: "credit\_card\_9513926"\\
\}\\
$\rightarrow$ \textbf{success}
\end{tcolorbox}

\begin{tcolorbox}[colback=red!3, colframe=tbred, boxrule=0.4pt, arc=2pt, left=6pt, right=6pt, top=4pt, bottom=4pt, fontupper=\small]
\textcolor{tbred}{\textbf{ReAct Failure Mode}}\\[3pt]
ReAct attempts to exchange for the clicky+RGB+full variant (item \texttt{9690244451}) without checking availability, causing an API error. After retrying, it \textbf{loses track of the conditional fallback} (``if no RGB, then no backlight'') and either:
\begin{itemize}[leftmargin=1.5em, itemsep=1pt, topsep=2pt]
\item selects a wrong variant (e.g., white backlight instead of none), or
\item only exchanges one of the two items, forgetting the thermostat.
\end{itemize}
Result: 2 to 3 additional correction turns required, task only partially completed.
\end{tcolorbox}

\end{tcolorbox}

\subsection{VitaBench Cross-domain: CS Trip with Schedule Conflict and Delivery Constraint}

\begin{tcolorbox}[
    colback=white,
    colframe=factframe,
    title={\textbf{Example 2: Cascading time constraints across train, CS booking, and delivery}},
    fonttitle=\small\sffamily\bfseries,
    coltitle=white,
    colbacktitle=factframe,
    boxrule=0.5pt, arc=3pt,
    left=8pt, right=8pt, top=6pt, bottom=6pt,
    fontupper=\small,
    breakable,
    pad at break=3pt
]

\noindent\textbf{Benchmark and task}: VitaBench Cross-domain, schedule repair across transportation, attraction booking, and delivery.\\[2pt]
\textbf{User goal}:\\[2pt]
\textit{User U624274 (female engineer, age 30--35, Baoding) and 5~friends plan a real-person CS activity in Beijing on Saturday. Existing orders: train Baoding$\rightarrow$Beijing (G505, Sat), return train (Sun), CS tickets (6~adult, Sat 14:00). Company announces a Saturday morning meeting, requiring later departure and cascading rescheduling.}

\tcbline

\textbf{Turn 1} \textcolor{promptframe}{\textbf{[User]}}\\
\textit{``My company scheduled a meeting Saturday morning. I need a high-speed rail departing after 1:00~PM; please ask my friend to rebook. The CS activity needs to be pushed back. Allow about 1.5h travel from station to CS base.''}

\tcbline

\textbf{Turn 2} \textcolor{factframe}{\textbf{[Tool Call]}} \texttt{search\_trains}\\[2pt]
\begin{tcolorbox}[colback=factbg, colframe=factframe!30, boxrule=0.3pt, arc=1pt, left=4pt, right=4pt, top=2pt, bottom=2pt, fontupper=\small\ttfamily]
search\_trains(\{from: "Baoding", to: "Beijing",\\
\hspace*{14em}date: "2025-06-21", depart\_after: "13:00"\})\\[4pt]
$\rightarrow$ [\\
~~\{ train\_id: "T00012",\\
~~~~number: "G661",\\
~~~~depart: "13:28",~~arrive: "14:15",\\
~~~~price: 95,~~seat\_type: "Second class" \},\\[2pt]
~~\{ train\_id: "T00014",\\
~~~~number: "G515",\\
~~~~depart: "14:02",~~arrive: "14:48",\\
~~~~price: 92,~~seat\_type: "Second class" \},\\
~~...\\
]
\end{tcolorbox}

\textbf{Turn 3} \textcolor{factframe}{\textbf{[Tool Call]}}\\
\texttt{get\_order("S17550802154486284\_O00004")}\\[2pt]
\begin{tcolorbox}[colback=factbg, colframe=factframe!30, boxrule=0.3pt, arc=1pt, left=4pt, right=4pt, top=2pt, bottom=2pt, fontupper=\small\ttfamily]
\{\\
~~order\_type: "attraction",\\
~~store: "Beijing Tactical CS Base, Huilongguan, Changping District",\\
~~status: "paid",\\
~~products: [\{ name: "Adult ticket", price: 398, qty: 6,\\
~~~~~~~~~~~~~~date: "2025-06-21" \}],\\
~~\textbf{reservation\_time: "14:00"},\\
~~location: \{ lat: 40.098765, lng: 116.345678 \}\\
\}
\end{tcolorbox}

\begin{tcolorbox}[colback=statebg, colframe=stateframe, boxrule=0.4pt, arc=2pt, left=4pt, right=4pt, top=3pt, bottom=3pt, fontupper=\footnotesize\ttfamily]
\textcolor{stateframe}{\textbf{IDSS State (After Turn 3)}}\\[2pt]
\textcolor{stateframe}{\textbf{Fact Layer:}}\\
~~[Entity: Train] id: G661\\
~~~~source: search\_trains()\\
~~~~depart: 13:28, arrive: 14:15, price: 95\\
~~[Entity: Train] id: G515\\
~~~~source: search\_trains()\\
~~~~depart: 14:02, arrive: 14:48, price: 92\\
~~[Entity: Reservation] id: CS\_O00004\\
~~~~source: get\_order()\\
~~~~venue: "Beijing Tactical CS Base"\\
~~~~reservation\_time: 14:00, pax: 6, status: paid\\[2pt]
\textcolor{stateframe}{\textbf{State Layer:}}\\[1pt]
~~\textcolor{stateframe}{\textit{Intents:}}\\
~~~~I1: "Select train after 13:00" $\rightarrow$ \textbf{active}\\
~~~~I2: "Reschedule CS reservation" $\rightarrow$ \textbf{pending}\\
~~~~~~\textit{depends\_on}: [I1.resolved]\\
~~~~I3: "Notify friend" $\rightarrow$ \textbf{pending}\\
~~~~~~\textit{depends\_on}: [I1.resolved]\\[1pt]
~~\textcolor{stateframe}{\textit{Variables:}}\\
\hspace*{1em}\textbf{departure\_constraint}: "after 13:00"\\
\hspace*{2em}source: askable | known: true\\
\hspace*{1em}\textbf{travel\_time\_to\_cs}: "1.5h"\\
\hspace*{2em}source: askable | known: true\\
\hspace*{1em}\textbf{selected\_train}: null\\
\hspace*{2em}source: derivable | known: false\\
\hspace*{1em}\textbf{new\_cs\_time}: null\\
\hspace*{2em}source: derivable | known: false\\[1pt]
~~\textcolor{stateframe}{\textit{Constraints:}}\\
\hspace*{1em}C1: "depart $\geq$ 13:00" $\rightarrow$ \textbf{satisfied}\\
\hspace*{2em}evidence: "G661=13:28, G515=14:02"\\
\hspace*{1em}C2: "arrive+travel $\leq$ CS start" $\rightarrow$ \textbf{violated}\\
\hspace*{2em}evidence: "14:15+1.5h=15:45 > 14:00"\\
\hspace*{1em}C3: "CS starts before 17:00" $\rightarrow$ \textbf{unsatisfied}\\
\hspace*{2em}evidence: "new\_cs\_time unknown"\\[1pt]
~~\textcolor{stateframe}{\textit{Next Action:}}\\
\hspace*{1em}Recommend G661 and ask for confirmation.
\end{tcolorbox}

\tcbline

\textbf{Turn 4} \textcolor{stateframe}{\textbf{[IDSS Agent]}}\\
\textit{``I found G661 departing at 13:28 (arriving 14:15). With 1.5h travel to the CS base, you'd arrive around 15:45. I'll reschedule the CS reservation from 14:00 to 15:30 for a buffer. G515 departs at 14:02, pushing CS start to 16:18---possibly too late for a full afternoon session. Shall I book G661 and update the reservation to 15:30?''}

\tcbline

\textbf{Turn 5} \textcolor{promptframe}{\textbf{[User]}}\\
\textit{``Yes, G661 is good. Also, I need sunscreen before the trip. Check if Anessa can be delivered to my home today before 10 PM.''}

\tcbline

\textbf{Turn 6} \textcolor{factframe}{\textbf{[Tool Call]}} \texttt{search\_delivery}\\[2pt]
\begin{tcolorbox}[colback=factbg, colframe=factframe!30, boxrule=0.3pt, arc=1pt, left=4pt, right=4pt, top=2pt, bottom=2pt, fontupper=\small\ttfamily]
search\_delivery(\{\\
~~product: "Anessa sunscreen",\\
~~address: "Room 502, Unit 2, Building 8,\\
~~~~~~~~~~~Oriental Garden, Baoding"\\
\})\\[4pt]
$\rightarrow$ \{\\
~~available: true,\\
~~product: "Shiseido Anessa SPF50+ PA++++ 90ml",\\
~~price: 249,\\
~~store: "JD Beauty Supermarket",\\
~~delivery\_slot: "today 20:00 to 22:00"\\
\}
\end{tcolorbox}

\begin{tcolorbox}[colback=statebg, colframe=stateframe, boxrule=0.4pt, arc=2pt, left=4pt, right=4pt, top=3pt, bottom=3pt, fontupper=\footnotesize\ttfamily]
\textcolor{stateframe}{\textbf{IDSS State (After Turn 6)}}\\[2pt]
\textcolor{stateframe}{\textbf{Fact Layer:}}\\
~~[Entity: Delivery] id: pending\_order\\
~~~~source: search\_delivery()\\
~~~~product: "Anessa SPF50+ 90ml", price: 249\\
~~~~store: "JD Beauty Supermarket"\\
~~~~slot: today 20:00--22:00\\[2pt]
\textcolor{stateframe}{\textbf{State Layer:}}\\[1pt]
~~\textcolor{stateframe}{\textit{Intents:}}\\
~~~~I1: "Book train G661" $\rightarrow$ \textbf{active}\\
~~~~I2: "Reschedule CS to 15:30" $\rightarrow$ \textbf{active}\\
~~~~I3: "Notify friend" $\rightarrow$ \textbf{pending}\\
~~~~I4: "Order sunscreen delivery" $\rightarrow$ \textbf{active}\\[1pt]
~~\textcolor{stateframe}{\textit{Variables:}}\\
\hspace*{1em}\textbf{delivery\_deadline}: "before 22:00"\\
\hspace*{2em}source: askable | known: true\\
\hspace*{1em}\textbf{quiet\_delivery}: null\\
\hspace*{2em}source: askable | known: false\\
\hspace*{2em}evidence: "slot 20:00--22:00, late evening"\\[1pt]
~~\textcolor{stateframe}{\textit{Constraints:}}\\
\hspace*{1em}C4: "delivery $\leq$ 22:00" $\rightarrow$ \textbf{satisfied}\\
\hspace*{2em}evidence: "slot ends 22:00"\\
\hspace*{1em}C5: "quiet delivery" $\rightarrow$ \textbf{unsatisfied}\\
\hspace*{2em}evidence: "slot 20:00--22:00"\\[1pt]
~~\textcolor{stateframe}{\textit{Next Action:}}\\
\hspace*{1em}Confirm quiet-delivery, then place order.
\end{tcolorbox}

\tcbline

\textbf{Turn 7} \textcolor{stateframe}{\textbf{[IDSS Agent]}}\\
\textit{``Anessa SPF50+ 90ml (\textyen 249) can be delivered today between 8 and 10 PM. Since it's a late evening delivery, would you like me to add a note to knock quietly or leave at door?''}

\tcbline

\begin{tcolorbox}[colback=red!3, colframe=tbred, boxrule=0.4pt, arc=2pt, left=6pt, right=6pt, top=4pt, bottom=4pt, fontupper=\small]
\textcolor{tbred}{\textbf{ReAct Failure Mode}}\\[3pt]
ReAct changes the train ticket but \textbf{fails to propagate the time constraint} to the CS reservation (the dependency chain \texttt{train\_arrival}$\,\rightarrow\,$\texttt{travel\_time}$\,\rightarrow\,$\texttt{CS\_start} is not tracked). Specifically:
\begin{itemize}[leftmargin=1.5em, itemsep=1pt, topsep=2pt]
\item Books G661 but leaves CS reservation at 14:00 (impossible to attend).
\item Orders sunscreen without verifying the delivery time constraint.
\item After 3 additional user correction turns, plan still has timing inconsistency.
\end{itemize}
\end{tcolorbox}

\end{tcolorbox}

\subsection{UserBench: Business Trip with Underspecified Preferences}

\begin{tcolorbox}[
    colback=white,
    colframe=stateframe,
    title={\textbf{Example 3: Eliciting implicit preferences for flights and restaurants}},
    fonttitle=\small\sffamily\bfseries,
    coltitle=white,
    colbacktitle=stateframe,
    boxrule=0.5pt, arc=3pt,
    left=8pt, right=8pt, top=6pt, bottom=6pt,
    fontupper=\small,
    breakable,
    pad at break=3pt
]

\noindent\textbf{Benchmark and task}: UserBench TravelGym, underspecified flight and restaurant preferences.\\[2pt]
\textbf{User goal}:\\[2pt]
\textit{``I am planning a business trip to NYC from December 10th to 15th. I want my flight from LA on December 10th to include carry-on baggage allowance~[...] Additionally, I need generous checked luggage service because I plan to bring back souvenirs. During my stay, dining in restaurants with a conducive environment for business discussions is essential. I usually avoid places with any 1- or 2-star reviews.''}\\[4pt]
\textbf{Hidden preferences} (ground truth, not revealed upfront):\\
~~\textbullet~\textit{Flight}: carry-on baggage allowance + checked bags service\\
~~\textbullet~\textit{Restaurant}: all reviews $\geq$ 3-star + ``Business Dining'' tag

\tcbline

\textbf{Turn 1} \textcolor{promptframe}{\textbf{[User]}}\\
\textit{``I need to book a flight from LA to NYC on Dec~10th and find two good restaurants for business dinners.''}

\tcbline

\textbf{Turn 2} \textcolor{stateframe}{\textbf{[IDSS Agent]}}\\
\textit{``I'd be happy to help. A few questions: (1)~Do you have luggage preferences---carry-on only, or do you need checked bags? (2)~For the restaurants, any cuisine preference? Is the ambiance important---casual, upscale, or specifically suited for business meetings?''}

\tcbline

\textbf{Turn 3} \textcolor{promptframe}{\textbf{[User]}}\\
\textit{``I always keep travel essentials with me for unexpected layovers, and I'm bringing back souvenirs so I'll need checked luggage too. For restaurants, the environment matters because the conversation might touch on business.''}

\tcbline

\begin{tcolorbox}[colback=statebg, colframe=stateframe, boxrule=0.4pt, arc=2pt, left=4pt, right=4pt, top=3pt, bottom=3pt, fontupper=\footnotesize\ttfamily]
\textcolor{stateframe}{\textbf{IDSS State (After Turn 3)}}\\[2pt]
\textcolor{stateframe}{\textbf{Fact Layer:}}~~(empty --- no tool calls yet)\\[2pt]
\textcolor{stateframe}{\textbf{State Layer:}}\\[1pt]
~~\textcolor{stateframe}{\textit{Intents:}}\\
~~~~I1: "Search flights" $\rightarrow$ \textbf{active}\\
~~~~~~\textit{reason}: flight preferences are known\\
~~~~I2: "Elicit rating preference" $\rightarrow$ \textbf{active}\\
~~~~~~\textit{reason}: rating threshold unknown\\
~~~~I3: "Search restaurants" $\rightarrow$ \textbf{pending}\\
~~~~~~\textit{depends\_on}: [I2.completed]\\[1pt]
~~\textcolor{stateframe}{\textit{Variables:}}\\
\hspace*{1em}\textbf{origin}: "LAX"\\
\hspace*{2em}source: askable | known: true\\
\hspace*{1em}\textbf{destination}: "NYC"\\
\hspace*{2em}source: askable | known: true\\
\hspace*{1em}\textbf{date}: "Dec 10"\\
\hspace*{2em}source: askable | known: true\\
\hspace*{1em}\textbf{carry\_on}: "required"\\
\hspace*{2em}source: derivable | known: true\\
\hspace*{2em}evidence: "keep travel essentials with me"\\
\hspace*{1em}\textbf{checked\_bags}: "required"\\
\hspace*{2em}source: derivable | known: true\\
\hspace*{2em}evidence: "bringing back souvenirs"\\
\hspace*{1em}\textbf{restaurant\_ambiance}: "business dining"\\
\hspace*{2em}source: derivable | known: true\\
\hspace*{2em}evidence: "conversation touches business"\\
\hspace*{1em}\textbf{rating\_threshold}: null\\
\hspace*{2em}source: askable | known: false\\[1pt]
~~\textcolor{stateframe}{\textit{Constraints:}}\\
\hspace*{1em}C1: "carry-on preference known" $\rightarrow$ \textbf{satisfied}\\
\hspace*{2em}evidence: "carry\_on = required"\\
\hspace*{1em}C2: "checked bags preference known" $\rightarrow$ \textbf{satisfied}\\
\hspace*{2em}evidence: "checked\_bags = required"\\
\hspace*{1em}C3: "dining ambiance known" $\rightarrow$ \textbf{satisfied}\\
\hspace*{2em}evidence: "ambiance = business dining"\\
\hspace*{1em}C4: "rating threshold known" $\rightarrow$ \textbf{unsatisfied}\\
\hspace*{2em}evidence: "rating\_threshold = null"\\[1pt]
~~\textcolor{stateframe}{\textit{Next Action:}}\\
\hspace*{1em}Ask about restaurant rating tolerance.
\end{tcolorbox}

\tcbline

\textbf{Turn 4} \textcolor{stateframe}{\textbf{[IDSS Agent]}}\\
\textit{``Got it---I'll filter for flights with both carry-on and checked bag service. One more thing about restaurants: do you have a minimum quality bar? For example, do you mind if a place has some lower ratings, or do you prefer consistently high reviews?''}

\tcbline

\textbf{Turn 5} \textcolor{promptframe}{\textbf{[User]}}\\
\textit{``I usually scroll past restaurants with any 1- or 2-star reviews. I don't want to take the risk, especially for a business dinner.''}

\tcbline

\begin{tcolorbox}[colback=statebg, colframe=stateframe, boxrule=0.4pt, arc=2pt, left=4pt, right=4pt, top=3pt, bottom=3pt, fontupper=\footnotesize\ttfamily]
\textcolor{stateframe}{\textbf{IDSS State (After Turn 5)}}\\[2pt]
\textcolor{stateframe}{\textbf{Fact Layer:}}~~(empty --- no tool calls yet)\\[2pt]
\textcolor{stateframe}{\textbf{State Layer:}}\\[1pt]
~~\textcolor{stateframe}{\textit{Intents:}}\\
~~~~I1: "Search flights" $\rightarrow$ \textbf{active}\\
~~~~I3: "Search restaurants" $\rightarrow$ \textbf{active}\\
~~~~~~\textit{depends\_on}: [I2.completed]\\[1pt]
~~\textcolor{stateframe}{\textit{Variables:}}\\
\hspace*{1em}\textbf{carry\_on}: "required"\\
\hspace*{2em}source: derivable | known: true\\
\hspace*{2em}evidence: "keep travel essentials with me"\\
\hspace*{1em}\textbf{checked\_bags}: "required"\\
\hspace*{2em}source: derivable | known: true\\
\hspace*{2em}evidence: "bringing back souvenirs"\\
\hspace*{1em}\textbf{restaurant\_ambiance}: "business dining"\\
\hspace*{2em}source: derivable | known: true\\
\hspace*{2em}evidence: "conversation touches business"\\
\hspace*{1em}\textbf{rating\_threshold}: "all reviews $\geq$ 3-star"\\
\hspace*{2em}source: askable | known: true\\
\hspace*{2em}evidence: "avoids 1--2 star reviews"\\[1pt]
~~\textcolor{stateframe}{\textit{Constraints:}}\\
\hspace*{1em}C1: "includes carry-on" $\rightarrow$ \textbf{unsatisfied}\\
\hspace*{1em}C2: "supports checked bags" $\rightarrow$ \textbf{unsatisfied}\\
\hspace*{1em}C3: "Business Dining tag" $\rightarrow$ \textbf{unsatisfied}\\
\hspace*{1em}C4: "no 1--2 star reviews" $\rightarrow$ \textbf{unsatisfied}\\
\hspace*{2em}\textit{(all pending tool verification)}\\[1pt]
~~\textcolor{stateframe}{\textit{Next Action:}}\\
\hspace*{1em}Search flights and restaurants.
\end{tcolorbox}

\tcbline

\textbf{Turn 6} \textcolor{factframe}{\textbf{[Tool Call]}} \texttt{interact\_with\_env} (flight search)\\[2pt]
\begin{tcolorbox}[colback=factbg, colframe=factframe!30, boxrule=0.3pt, arc=1pt, left=4pt, right=4pt, top=2pt, bottom=2pt, fontupper=\small\ttfamily]
search("flight LAX to NYC Dec 10") $\rightarrow$ [\\[2pt]
~~F8: \{ company: "Delta Airlines",\\
~~~~~~~~flight\_number: "DL456",\\
~~~~~~~~cost: \$350,\\
~~~~~~~~amenities: ["WiFi", "Carry on Baggage Allowance"],\\
~~~~~~~~service: \{ checked\_bag\_cost: \$50 \},\\
~~~~~~~~duration: 5.5h \},\\[3pt]
~~F16: \{ company: "American Airlines",\\
~~~~~~~~~flight\_number: "AA789",\\
~~~~~~~~~cost: \$400,\\
~~~~~~~~~amenities: ["WiFi", "Carry on Baggage Allowance"],\\
~~~~~~~~~service: \{ checked\_bag\_cost: \$60 \} \},\\[3pt]
~~F21: \{ company: "Southwest Airlines", ... \},\\
~~...\\
]
\end{tcolorbox}

\textbf{Turn 7} \textcolor{factframe}{\textbf{[Tool Call]}} \texttt{interact\_with\_env} (restaurant search)\\[2pt]
\begin{tcolorbox}[colback=factbg, colframe=factframe!30, boxrule=0.3pt, arc=1pt, left=4pt, right=4pt, top=2pt, bottom=2pt, fontupper=\small\ttfamily]
search("restaurant NYC Dec 11") $\rightarrow$ [\\[2pt]
~~R17: \{ name: "The Executive Table",\\
~~~~~~~~~cuisine: "French",\\
~~~~~~~~~rating: 9,\\
~~~~~~~~~reviews: \{ 5\,\(\star\): 320, 4\,\(\star\): 180, 3\,\(\star\): 50,\\
~~~~~~~~~~~~~~~~~~~\textbf{2\,\(\star\): 0, 1\,\(\star\): 0} \},\\
~~~~~~~~~tags: ["Business Dining", "Reservations", "Parking"] \},\\[3pt]
~~R8:~~\{ name: "Sushi Prestige",\\
~~~~~~~~~cuisine: "Japanese",\\
~~~~~~~~~rating: 8,\\
~~~~~~~~~reviews: \{ ..., \textbf{2\,\(\star\): 0, 1\,\(\star\): 0} \},\\
~~~~~~~~~tags: ["Business Dining", "Walk-ins Welcomed"] \},\\[3pt]
~~R21: \{ name: "Urban Grill",\\
~~~~~~~~~reviews: \{ ..., \textbf{2\,\(\star\): 5, 1\,\(\star\): 2} \},\\
~~~~~~~~~tags: ["Casual Dining", "Late Night"] \},\\
~~...\\
]
\end{tcolorbox}

\tcbline

\textbf{Turn 8} \textcolor{stateframe}{\textbf{[IDSS Agent]}}\\
\textit{``Here are my recommendations:}\\[2pt]
\textit{\textbf{Flight:} Delta DL456 (\$350 + \$50 checked bag), including carry-on allowance and checked bag service.}\\[2pt]
\textit{\textbf{Restaurants:}}\\
\textit{~~(1) The Executive Table (French, all reviews 3-star+, Business Dining tag)}\\
\textit{~~(2) Sushi Prestige (Japanese, all reviews 3-star+, Business Dining tag)}\\[2pt]
\textit{I filtered out places like Urban Grill that have 1- to 2-star reviews. Shall I book these?''}

\begin{tcolorbox}[colback=statebg, colframe=stateframe, boxrule=0.4pt, arc=2pt, left=4pt, right=4pt, top=3pt, bottom=3pt, fontupper=\footnotesize\ttfamily]
\textcolor{stateframe}{\textbf{IDSS State (After Turn 7)}}\\[2pt]
\textcolor{stateframe}{\textbf{Fact Layer:}}\\
~~[Entity: Flight] id: DL456\\
~~~~source: interact\_with\_env()\\
~~~~company: Delta, cost: \$350\\
~~~~amenities: [WiFi, Carry on Baggage Allowance]\\
~~~~checked\_bag\_cost: \$50\\
~~[Entity: Restaurant] id: R17\\
~~~~source: interact\_with\_env()\\
~~~~name: "The Executive Table", cuisine: French\\
~~~~min\_review: 3$\star$, tags: [Business Dining]\\
~~[Entity: Restaurant] id: R8\\
~~~~source: interact\_with\_env()\\
~~~~name: "Sushi Prestige", cuisine: Japanese\\
~~~~min\_review: 3$\star$, tags: [Business Dining]\\
~~\textcolor{tbred}{[Entity: Restaurant] id: R21 --- \textbf{filtered}}\\
~~~~\textcolor{tbred}{min\_review: 1$\star$ $\rightarrow$ violates C4}\\[2pt]
\textcolor{stateframe}{\textbf{State Layer:}}\\[1pt]
~~\textcolor{stateframe}{\textit{Intents:}}\\
~~~~I1: "Book flight" $\rightarrow$ \textbf{completed}\\
~~~~I3: "Recommend restaurants" $\rightarrow$ \textbf{completed}\\[1pt]
~~\textcolor{stateframe}{\textit{Variables:}}\\
\hspace*{1em}\textbf{carry\_on}: "required"\\
\hspace*{2em}source: derivable | known: true\\
\hspace*{1em}\textbf{checked\_bags}: "required"\\
\hspace*{2em}source: derivable | known: true\\
\hspace*{1em}\textbf{rating\_threshold}: "all reviews $\geq$ 3-star"\\
\hspace*{2em}source: askable | known: true\\[1pt]
~~\textcolor{stateframe}{\textit{Constraints:}}\\
\hspace*{1em}C1: "includes carry-on" $\rightarrow$ \textbf{satisfied}\\
\hspace*{2em}evidence: "DL456 amenities"\\
\hspace*{1em}C2: "supports checked bags" $\rightarrow$ \textbf{satisfied}\\
\hspace*{2em}evidence: "DL456 checked\_bag\_cost=\$50"\\
\hspace*{1em}C3: "Business Dining tag" $\rightarrow$ \textbf{satisfied}\\
\hspace*{2em}evidence: "R17, R8 tagged"\\
\hspace*{1em}C4: "no 1--2 star reviews" $\rightarrow$ \textbf{satisfied}\\
\hspace*{2em}evidence: "R17 min=3 star, R8 min=3 star;\\
\hspace*{2em}R21 has 2 star $\rightarrow$ excluded"\\[1pt]
~~\textcolor{stateframe}{\textit{Next Action:}}\\
\hspace*{1em}Present recommendations, await confirmation.
\end{tcolorbox}

\tcbline

\begin{tcolorbox}[colback=red!3, colframe=tbred, boxrule=0.4pt, arc=2pt, left=6pt, right=6pt, top=4pt, bottom=4pt, fontupper=\small]
\textcolor{tbred}{\textbf{ReAct Failure Mode}}\\[3pt]
ReAct immediately searches flights without asking about baggage needs:
\begin{itemize}[leftmargin=1.5em, itemsep=1pt, topsep=2pt]
\item Selects cheapest flight \textbf{without checked bag service} (user preference missed).
\item For restaurants, picks highest-rated option without filtering for ``no 1--2 star reviews'' or ``Business Dining'' tag; recommends Urban Grill (has 2-star reviews).
\item \textbf{Never elicits} user's implicit preferences about baggage or review threshold.
\end{itemize}
Result: Low Preference Elicitation (PE) score. The user receives misaligned recommendations.
\end{tcolorbox}

\end{tcolorbox}

\end{document}